\documentclass{article}

\usepackage{PRIMEarxiv}

\usepackage[utf8]{inputenc} 
\usepackage[T1]{fontenc}    
\usepackage{hyperref}       
\usepackage{url}            
\usepackage{booktabs}       
\usepackage{amsfonts}       
\usepackage{nicefrac}       
\usepackage{microtype}      
\usepackage{lipsum}
\usepackage{fancyhdr}       
\usepackage{graphicx}       
\graphicspath{{media/}}     

\usepackage{soul}
\usepackage{url}
\usepackage{graphicx}
\usepackage{amsmath}
\usepackage{amsthm}
\usepackage{amsfonts}
\usepackage{amssymb}
\usepackage{mathtools}
\usepackage{booktabs}
\usepackage{algorithm}
\usepackage{algorithmic}
\usepackage{dsfont}
\usepackage{booktabs}       
\usepackage{nicefrac}       
\usepackage{microtype}      
\usepackage{xcolor}         

\usepackage{placeins}
\usepackage{tikz}

\newcommand{\R}{\mathbb{R}}
\newcommand{\Z}{\mathbb{Z}}

\definecolor{c1}{RGB}{161, 54, 112}
\definecolor{c2}{RGB}{236, 65, 118}
\definecolor{c3}{RGB}{154, 119, 207}
\definecolor{c4}{RGB}{84, 56, 132}

\newcommand{\DrawBox}[7]{
    \begin{scope}[line join=round, fill opacity=0.5, thick]
        \draw[fill=#7]  (#1,#2,#3) -- (#1,#2,#3+#6) -- (#1,#2+#5,#3+#6) -- (#1,#2+#5,#3) -- cycle;
        \draw[fill=#7]  (#1,#2,#3) -- (#1+#4,#2,#3) -- (#1+#4,#2+#5,#3) -- (#1,#2+#5,#3) -- cycle;
        \draw[fill=#7]  (#1,#2+#5,#3) -- (#1+#4,#2+#5,#3) -- (#1+#4,#2+#5,#3+#6) -- (#1,#2+#5,#3+#6) -- cycle;
        \draw[fill=#7]  (#1+#4,#2,#3) -- (#1+#4,#2,#3+#6) -- (#1+#4,#2+#5,#3+#6) -- (#1+#4,#2+#5,#3) -- cycle;
        \draw[fill=#7]  (#1,#2,#3+#6) -- (#1+#4,#2,#3+#6) -- (#1+#4,#2+#5,#3+#6) -- (#1,#2+#5,#3+#6) -- cycle;
        \draw[fill=#7]  (#1,#2,#3) -- (#1+#4,#2,#3) -- (#1+#4,#2,#3+#6) -- (#1,#2,#3+#6) -- cycle;
    \end{scope}
}

\newtheorem{proposition}{Proposition}

\title{Scale Equivariant U-Net}

\author{
  Mateus Sangalli, Samy Blusseau, Santiago Velasco-Forero, Jesús Angulo \\
  Center for Mathematical Morphology \\
  Mines ParisTech, PSL University \\
  Fontainebleau, France\\
  \texttt{\{mateus.sangalli, sammy.blusseau, santiago.velasco, jesus.angulo\}@minesparis.psl.eu} \\
}

\begin{document}

\maketitle

\begin{abstract}
In neural networks, the property of being equivariant to transformations improves generalization when the corresponding symmetry is present in the data. In particular, scale-equivariant networks are suited to computer vision tasks where the same classes of objects appear at different scales, like in most semantic segmentation tasks. Recently, convolutional layers equivariant to a semigroup of scalings and translations have been proposed. However, the equivariance of subsampling and upsampling has never been explicitly studied even though they are necessary building blocks in some segmentation architectures. The U-Net is a representative example of such architectures, which includes the basic elements used for state-of-the-art semantic segmentation. Therefore, this paper introduces the Scale Equivariant U-Net (SEU-Net), a U-Net that is made approximately equivariant to a semigroup of scales and translations through careful application of subsampling and upsampling layers and the use of aforementioned scale-equivariant layers. Moreover, a scale-dropout is proposed in order to improve generalization to different scales in approximately scale-equivariant architectures. The proposed SEU-Net is trained for semantic segmentation of the Oxford Pet IIIT and the DIC-C2DH-HeLa dataset for cell segmentation. The generalization metric to unseen scales is dramatically improved in comparison to the U-Net, even when the U-Net is trained with scale jittering, and to a scale-equivariant architecture that does not perform upsampling operators inside the equivariant pipeline. The scale-dropout induces better generalization on the scale-equivariant models in the Pet experiment, but not on the cell segmentation experiment.
\end{abstract}

\section{Introduction}
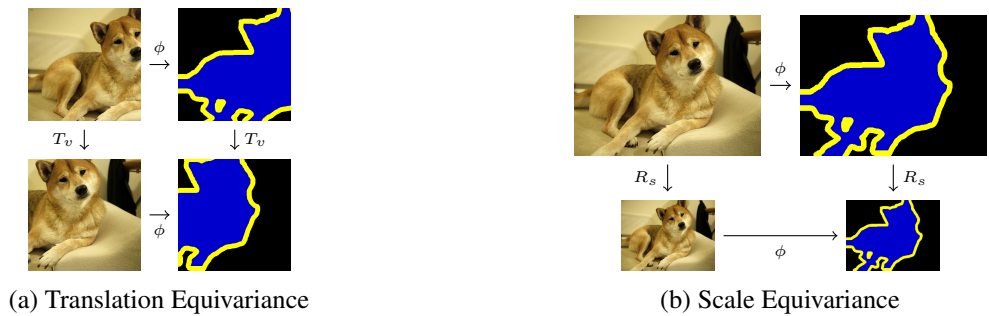
\begin{figure}[t]
    \centering
    \begin{minipage}[b]{.5\linewidth}
    \centering
        \begin{tikzpicture}[node distance=\textwidth, auto]
   
        \node (in) at (0,2) {\includegraphics[width=.18\textwidth]{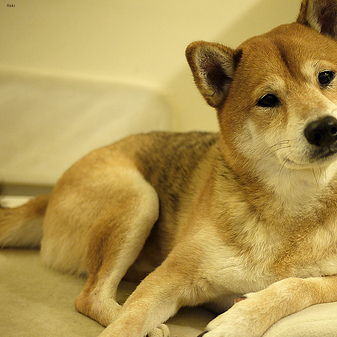}} ;
        \node (translate) at (0,0) {\includegraphics[width=.18\textwidth]{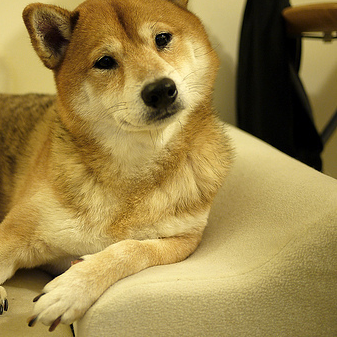}} ;
        \node (convolve) at (2,2) {\includegraphics[width=.18\textwidth]{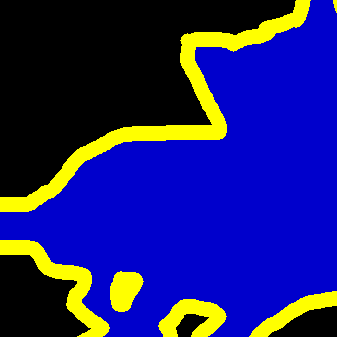}} ;
        \node (transconv) at (2,0) {\includegraphics[width=.18\textwidth]{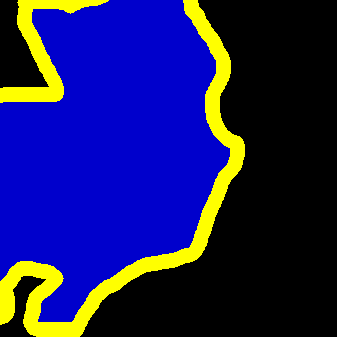}} ;
        
        \draw [->] (in) to node [left] {\tiny $T_v$} (translate) ;
        \draw [->] (in) to node [above] {\tiny $\phi$} (convolve) ;
        \draw [->] (translate) to node [below] {\tiny $\phi$} (transconv) ;
        \draw [->] (convolve) to node {\tiny $T_v$} (transconv) ;
   
        \end{tikzpicture}
        
        (a) Translation Equivariance
    \end{minipage}%
    \begin{minipage}[b]{.5\linewidth}
    \centering
        \begin{tikzpicture}[node distance=\textwidth, auto]

        \node (in) at (0,2) {\includegraphics[width=.3\textwidth]{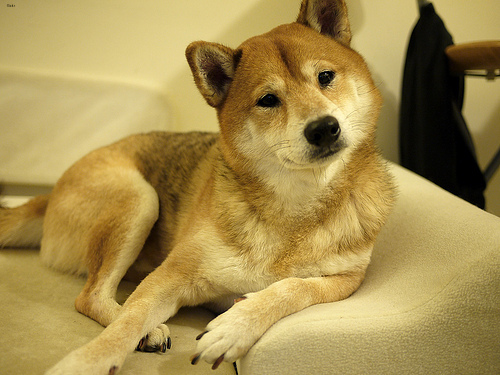}} ;
        \node (translate) at (0,0) {\includegraphics[width=.15\textwidth]{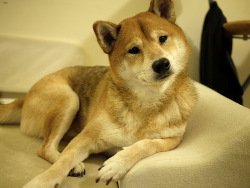}} ;
        \node (convolve) at (3,2) {\includegraphics[width=.3\textwidth]{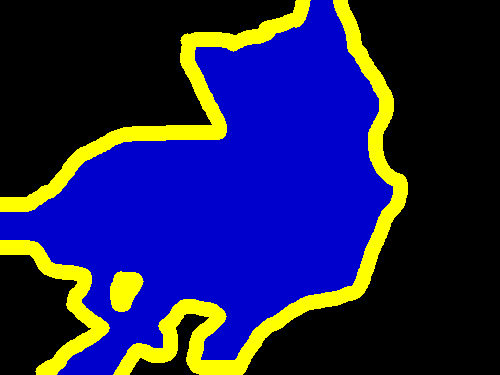}} ;
        \node (transconv) at (3,0) {\includegraphics[width=.15\textwidth]{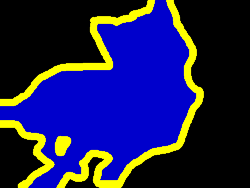}} ;
        
        \draw [->] (in) to node [left] {\tiny $R_s$} (translate) ;
        \draw [->] (in) to node [above] {\tiny $\phi$} (convolve) ;
        \draw [->] (translate) to node [below] {\tiny $\phi$} (transconv) ;
        \draw [->] (convolve) to node {\tiny $R_s$} (transconv) ;

        \end{tikzpicture}
    
        (b) Scale Equivariance
    \end{minipage}
    \caption{Example of equivariance in the cases of translation and scaling. In this case, $\phi$ is an ideal operator that computes the semantic segmentation of images. The operators $T_v$ and $R_s$ are, respectively, a translation and a re-scaling.}
    \label{fig:examples_equivariance}
\end{figure}

Convolutional Neural Networks (CNN) are based on convolutional layers and achieve state-of-the-art performance in many image analysis tasks. A translation applied to the inputs of a CNN is equivalent to a translation applied to its features maps, a property illustrated by Figure \ref{fig:examples_equivariance}(a). This property is a particular case of group equivariance \cite{cohen2016group} and helps improve the generalization of the network to new data if the data has translation symmetry. An operator $\phi:\mathcal{X} \to \mathcal{Y}$ is equivariant w.r.t. a group if applying a group action in the input and then $\phi$, amounts applying a group action to the output of  $\phi$ given the original inputs. This is illustrated in Figure \ref{fig:examples_equivariance}. In addition to translations, group actions can model many interesting classes of spatial transformations such as rotations, scalings, and affine transformations. 
Group equivariant CNNs~\cite{cohen2016group} are a generalization of CNNs that are equivariant to some transformation group. Many approaches focus on equivariance to rotations, in different kinds of data \cite{cohen2016group,weiler2018steerable,weiler20183d,thomas2018tensor} and to scalings \cite{zhu2019scale,ghosh2019scale,lindeberg2021scale}.

Deep scale-spaces \cite{worrall2019scale} introduce neural networks equivariant to the action of semigroups, instead of groups. Semigroup actions are considered as they can model non-invertible transformations, and the authors focus on equivariance to downsampling in discrete domains as a way to address equivariance to scalings without creating spurious information through interpolation.
This seminal work laid the basis to define scale-equivariant CNNs, although it only focused on convolutional layers and did not address the equivariance of pooling and upsampling layers, which are key elements in many neural architectures, such as U-Net.

\begin{figure}[t]
    \centering
    \begin{minipage}{.3\linewidth}
        \centering
        \includegraphics[width=.3\linewidth]{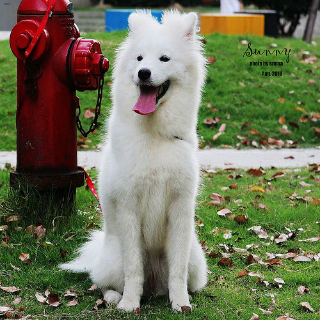}
        \includegraphics[width=.3\linewidth]{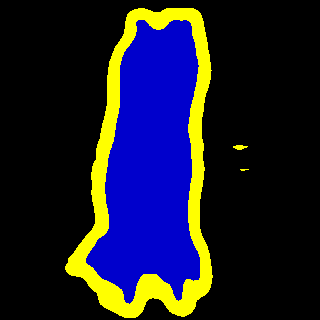}
        
        (a) Training scale
    \end{minipage}
    \begin{minipage}{.6\linewidth}
        \centering
        \includegraphics[width=.3\linewidth]{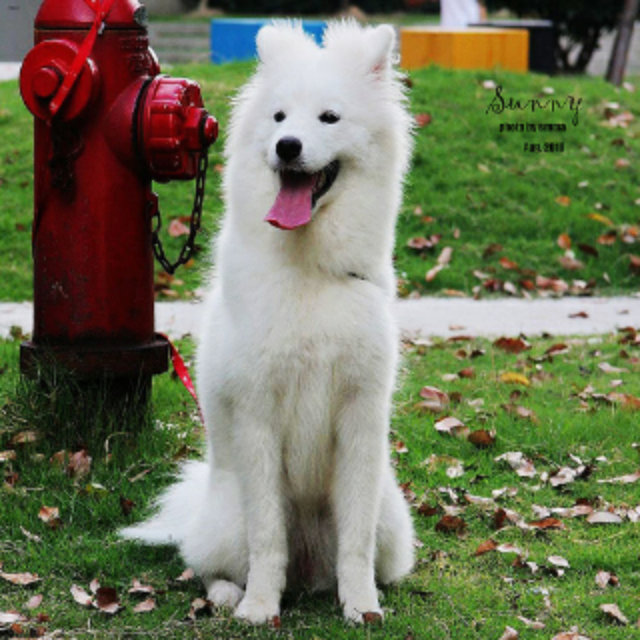}
        \includegraphics[width=.3\linewidth]{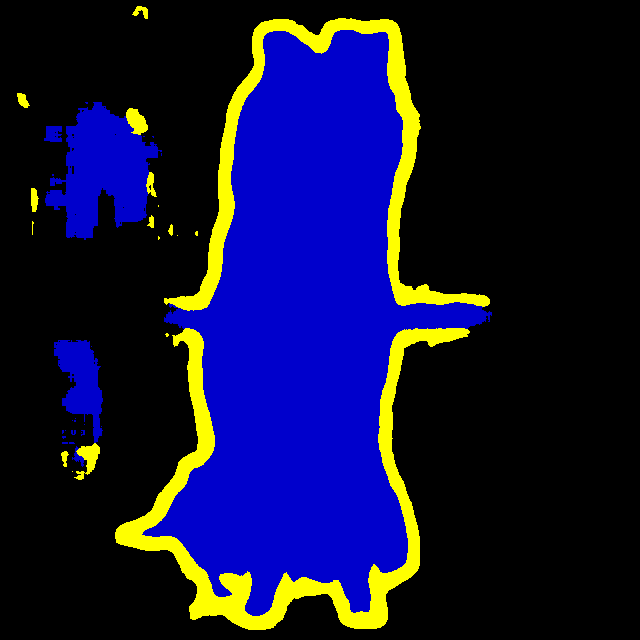}
        
        (b) Unseen scale
    \end{minipage}
    \caption{Example where a U-Net trained on one scale and is applied to predict an output on the training(a) and an unseen(b) scale. The image with the unseen scale represents the same object but the U-Net no longer segments it correctly.}
    \label{fig:example_no_equivariance}
\end{figure}

The U-Net~\cite{ronneberger2015unet} has become famous for its great performance in semantic segmentation.
It is a fully convolutional neural network, \emph{i.e.} a CNN without any dense layer, and therefore it is equivariant to a certain subgroup of translations. 
However, architectures like U-Net are not scale equivariant \textit{a priori}, and experiments show they are not in practice \cite{sangalli2021scale} as illustrated by Figure~\ref{fig:example_no_equivariance}. A scale-equivariant counterpart of such an architecture is desirable as scale symmetry is frequently present in semantic segmentation data.
For example, in urban scenes, objects of the same class appear at different scales depending on their distances to the camera.

In this work we introduce the Scale-Equivariant U-Net (SEU-Net) based on semigroup cross-correlations~\cite{worrall2019scale} and an adapted use of pooling and upsampling.
The rest of the paper is organized as follows. In Section \ref{sec:related} we discuss some of the related work in the literature. In Section \ref{sec:semigroup_networks} we review the semigroup equivariant neural networks. The main contribution of this paper, the SEU-Net, is introduced in Section \ref{sec:scale-unet} along with its fundamental building blocks.
The whole architecture is tested empirically for its equivariance in Section \ref{sec:experiments}.
More precisely, we test the SEU-Net \footnote{Code available at \url{https://github.com/mateussangalli/ScaleEquivariantUNet}} in segmentation tasks where the test images are in scales unseen during training, on the Oxford-IIIT Pet \cite{parkhi12a} and the DIC-HeLa cell \cite{ulman2017objective} datasets.
The SEU-Net is shown to overperform the U-Net even when the latter is training with large values of scale jittering.
The paper ends in Section \ref{sec:conclusions} with some conclusions and perspectives for future work.

\section{Related Work}
\label{sec:related}
Scale-equivariance and scale-invariance are topics already discussed in the deep learning literature \cite{zhu2019scale,ghosh2019scale,jansson2020unseenscales,lindeberg2021scale,sosnovik2019scale}.
The experimental benchmarks found in those papers are interesting as a first way to measure equivariance, but tend to be based on very simple tasks, such as the classification of re-scaled digits from the MNIST dataset or low resolution images of clothes from the Fashion-MNIST dataset. 
In \cite{sosnovik2021disco}, combinations of base filters are optimized to minimize the equivariance error of discrete scale convolutions. This is applied to classification, tracking and geometry estimation, but not segmentation.

In \cite{worrall2019scale}, instead of treating the scaling as an invertible operation, such as it would behave in a continuous domain, it is considered the action of downsampling the input image in a discrete domain. 
Therefore a \emph{semigroup}-equivariant generalization of the convolution is introduced. Specifically, the focus is put on a semigroup of scalings and translations. These operators can be efficiently applied even on large images, since applying it at larger scales has the same computational cost. In \cite{worrall2019scale} the semigroup equivariant models were applied to classification and semantic segmentation of datasets of large images, achieving
better results compared to matched non-equivariant architectures. Yet, the role of scale-equivariance was not isolated, as the performance of the models was not measured for inputs on scales unseen in the training set.
Later on, this approach was revisited by \cite{sangalli2021scale}, where the Gaussian scale-space originally used was generalized to other scale-spaces and the models were tested in experiments where the networks are trained in one fixed scale and tested on unseen scales, albeit on synthetic or simple datasets. In all these approaches, the authors either avoided pooling and upsampling in their architectures, or used them but did not discuss their impact on scale equivariance.

While scale-equivariance has been a topic in the literature for some time, as far as we know a scale-equivariant U-Net has not yet been proposed, contrary to the rotation-equivariance case \cite{chidester2019enhanced}.
Moreover, the current benchmarks for scale-equivariance were either based on simple datasets like MNIST or did not explicitly measure the equivariance in their segmentation or classification experiments, by training the networks on one fixed scale and testing on unseen scales.
Here we propose semantic segmentation experiments based on natural data which measure the equivariance of the predictions.

\section{Semigroup Equivariant Convolutional Networks}
\label{sec:semigroup_networks} 
In this work and following~\cite{worrall2019scale}, image scalings are restricted to image downscalings, which can be viewed as actions of a \emph{semigroup} on images. As illustrated by Figure~\ref{fig:examples_equivariance}, we seek equivariance with respect to both downscalings and translations. Hence, the network layers are designed to be equivariant with respect to a semigroup combining both transformations.

\subsection{Semigroup Equivariance}
\label{sec:semigroup-equivariance}

A semigroup, contrary to a group, can model non-invertible transformations, e.g. the downsampling operation in a discrete domain. In the following,  $(G, \cdot)$ denotes a discrete semigroup.

Let $\mathcal{X}$ be a set, a family of mappings $(\varphi_g)_{g\in G}$ from $\mathcal{X}$ to itself,  is a semigroup action on $\mathcal{X}$ if it is homomorphic to the semigroup, that is, if either
$\forall g, h \in G$,
$\varphi_g \circ \varphi_h = \varphi_{g\cdot h}$ (left action),
or
$\forall g, h \in G$,
$\varphi_g \circ \varphi_h = \varphi_{h\cdot g}$ (right action). In this paper we will consider in particular the following right action, acting on $\mathcal{F}$ the set of functions from $G$ to $\R^n$.
\begin{equation}
\label{eq:right-action}
\forall u,g\in G, \forall f\in\mathcal{F},\;\;\; R_u(f)(g) = f(u \cdot g).
\end{equation}

Given two sets $\mathcal{X}$ and $\mathcal{Y}$, a mapping $H:\mathcal{X}\to \mathcal{Y}$ is said equivariant with respect to $G$ if there are semigroup actions $(\varphi_g)_{g\in G}$ and $(\psi_g)_{g\in G}$ on $\mathcal{X}$ and $\mathcal{Y}$ respectively, such that $\forall g \in G, \;\;\; H\circ \varphi_g = \psi_g\circ H$.
This definition gets more intuitive when $\mathcal{X}=\mathcal{Y}$ is a set of images and $(\varphi_g)_{g\in G} = (\psi_g)_{g\in G}$ are scalings or translations, as illustrated in Figure~\ref{fig:examples_equivariance}.


\subsection{Scale-cross-correlation}
For $\gamma > 1$ an integer, let
$\mathcal{S}_\gamma = \{\gamma^n | n \in \mathbb{N} \}$, endowed with the multiplication, the semigroup representing discrete scalings of base $\gamma$. Then we consider the semigroup $G = \mathcal{S}_\gamma \times \Z^2$ of discrete scalings and translations, endowed with the internal operator $``\cdot"$, defined by
\begin{equation}
    \label{eq:composition-discrete-sg-scalings-translations}
    \forall k, l\in\mathbb{N}, z,y \in\mathbb{Z}^2 \;\;\; (\gamma^k,z)\cdot (\gamma^l, y) = (\gamma^{k+l}, \gamma^k y + z).
\end{equation}
Following \eqref{eq:right-action}, the action of this semigroup on functions mapping $\mathcal{S}_\gamma \times \mathbb{Z}^2$  to $\mathbb{R}$ is $R_{\gamma^k, z}[f](\gamma^l,y) = f(\gamma^{k+l}, \gamma^k y + z)$.
In analogy to convolutions, which are linear and equivariant to translations, a key step in  equivariant CNNs is defining linear operators which are equivariant to some class of operators.
The \emph{semigroup cross-correlation}, defined for an image $f:G \to \R$ and a filter $h:G \to \R$ is a generalization of the convolution which is linear and equivariant to the action $R_g$ of a semigroup.
When applied to the semigroup of scales and translations, we obtain the \emph{scale-cross-correlation}.
Both were introduced in \cite{worrall2019scale}. The scale-cross-correlation is written\footnote{The equations in the case of a general semigroup can be found in Appendix A.}
\begin{equation} \label{eq:scale_semigroup_corr_new}
    (f \star_G h)(\gamma^{k},z) = \sum\limits_{(\gamma^l, y) \in G} R_{\gamma^k, z}[f](\gamma^l, y) h(\gamma^l, y) 
     = \sum\limits_{l \geq 0} \sum\limits_{y \in \Z^2} f(\gamma^{k+l}, \gamma^k y + z) h(\gamma^l, y).
\end{equation}
This operator is suited for single channel images on $G$, but it can be easily extended to multichannel images. Let the input $f = (f_1, \dots, f_n) \in (\R^n)^G$ be a signal with $n$ channels. Assuming the output has $m$ channels, the filter is of the form $h: G \to \R^{n \times m}$. We compute the operator $f \star_G h$ at channel $o \in \{1, \dots, m\}$ as  $(f \star_G h)_o \coloneqq \sum\limits_{c=1}^n (f_c \star_G h_{c, o})$.
The resulting map is equivariant to scalings and translations: $(R_g f \star_G h)_o = R_g((f \star_G h)_o)$. Note that the composition of operators which commute with $R_g$ still commutes with $R_g$, for which concatenating scale-cross-correlation layers followed by pointwise activation functions and batch normalization yields equivariant architectures.


\subsection{Lifting and Projection}
The operators of the previous section are defined on the set of functions with the semigroup as a domain, $\mathcal{F} = (\R^n)^{G}$, but images input to networks are functions $f:\Z^2 \to \R^n$.
In this section we review the lifting and projections layers - operators which map images to functions on the semigroup and vice-versa.

\textbf{Lifting.}
A \emph{lifting} operator $\Lambda$ is used to map an input function $f:\Z^2 \to \R^n$ into a function $\Lambda f:G \to \R^n$. Once lifted to the semigroup space, linear equivariant operators can be applied according to Eq.~\eqref{eq:scale_semigroup_corr_new}.
As pointed out in~\cite{sangalli2021scale}, a sufficient condition to have equivariance of the composition of the lifting followed by the semigroup cross-correlations, is that $\Lambda \circ R^\prime_{\gamma^k, z} = R_{\gamma^k, z} \circ \Lambda$, where $R^\prime_{\gamma^k, z}$ is the re-scaling action for images on $\Z^2$: $R^\prime_{\gamma^k, z}[f](t) = f(\gamma^k t + z)$.


Whereas in \cite{sangalli2021scale} several liftings are explored, in this paper we set the lifting to the Gaussian scale-space $\Lambda_\mathcal{G}$, like in \cite{worrall2019scale}.
For an image $f: \Z^2 \to \R$ a point $z \in \Z^2$ and a scale level $k\in\mathbb{N}$,  
\begin{equation}
\label{q:gaussian-lifting}
    \Lambda_\mathcal{G}(f)(\gamma^k, z) = (f * {\mathcal{G}}_{\gamma^k})(z)
\end{equation}
where $*$ is the classic discrete convolution and ${\mathcal{G}}_{\gamma^k}$ the discrete  Gaussian kernel with scale $\gamma^k$. 

\textbf{Projection.}
To project back into the image space, we apply a \emph{max-projection} along the scale dimension, defined by $\forall z \in \Z^2 \; \Pi [f](z) = \sup_{k\in\mathbb{N}} \{f(\gamma^k,z)\}$.
To be consistent with the lifting, we would like to have
 $R^\prime_{\gamma^k, z} \circ\Pi = \Pi \circ R_{\gamma^k, z}$. Instead, we have $R^\prime_{\gamma^k, z} \Pi f(y) = \sup_{l\in\mathbb{N}} f (\gamma^l, \gamma^k y + z)$ and $\Pi R_{\gamma^k, z} f(y) = \sup_{l\in\mathbb{N}} f (\gamma^{l+k}, \gamma^k y + z) = \sup_{l\geq k} f (\gamma^l, \gamma^k y + z) $ so that $ R^\prime_{\gamma^k, z} \Pi f(y) = \max \{ \Pi R_{\gamma^k, z} f(y), \max_{0\leq l < k} f (\gamma^l, \gamma^k y+z) \}$.
The previous expression will be equivariant if the scale where the maximum is attained is smaller than $k$, but in general we can only hope for approximate equivariance for small enough $k$. The approximate equivariance will be empirically verified in experiments in Section \ref{sec:experiments}. Note that other projections (e.g. sum or average) have the same flaw, as this is intrinsic to the semigroup-equivariant approach, even though it was omitted previously in the literature.

\section{Scale-Equivariant U-Net}
\label{sec:scale-unet}

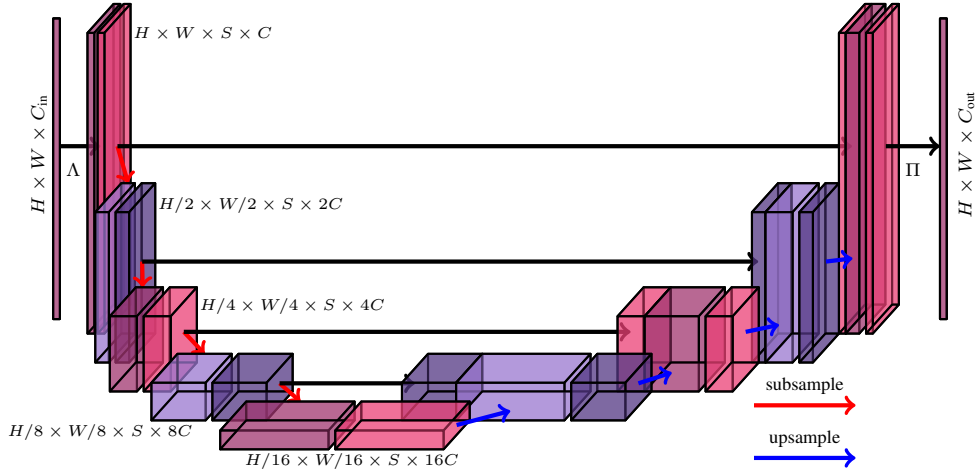
\begin{figure}
    \centering
    \begin{tikzpicture}[xscale=0.9]
    
    \DrawBox{-.7}{0}{.5}{.1}{4}{0}{c1}
    \draw[->, ultra thick, draw=black] (-.6, 2.3, .5) -- (0., 2.3, .5);
    \node[anchor=north] at (-.4, 2.2, .5) {\scriptsize $\Lambda$};
    \node at (-1.1, 2) {\scriptsize \rotatebox{90}{$H \times W \times C_\text{in}$}};
    
    \DrawBox{12.4}{0}{.5}{.1}{4}{0}{c1}
    \DrawBox{0  }{0}{0}{.1}{4}{1}{c1}
    \DrawBox{.15 }{0}{0}{.1}{4}{1}{c2}
    \draw[->, ultra thick, draw=red] (.25, 2.3, .5) -- (.6, 2, 1);
    \draw[->, ultra thick, draw=black] (.25, 2.3, .5) -- (11.1, 2.3, .5);
    \node[anchor=west] at (.35, 3.8, .5) {\tiny $H \times W \times S \times C$};
    
    \DrawBox{11.1}{0}{0}{.1}{4}{1}{c2}
    \DrawBox{11.2}{0}{0}{.2}{4}{1}{c1}
    \DrawBox{11.5}{0}{0}{.1}{4}{1}{c2}
    \draw[->, ultra thick, draw=black] (11.6, 2.3, .5) -- (12.4, 2.3, .5);
    \node[anchor=north] at (12, 2.2, .5) {\scriptsize $\Pi$};
    \node at (12.6, 2) {\scriptsize \rotatebox{90}{$H \times W \times C_\text{out}$}};
    
    \DrawBox{.5}{0}{1}{.2}{2}{1}{c3}
    \DrawBox{.8}{0}{1}{.2}{2}{1}{c4}
    \draw[->, ultra thick, draw=red] (1, 1.15, 1.5) -- (1.2, 1, 2);
    \draw[->, ultra thick, draw=black] (1, 1.15, 1.5) -- (10.1, 1.15, 1.5);
    \node[anchor=west] at (1.1, 1.9, 1.5) {\tiny $H/2 \times W/2 \times S \times 2C$};
    
    \DrawBox{10.2}{0}{1}{.2}{2}{1}{c4}
    \DrawBox{10.4}{0}{1}{.4}{2}{1}{c3}
    \DrawBox{10.9}{0}{1}{.2}{2}{1}{c4}
    \draw[->, ultra thick, draw=blue] (11.1, 1.15, 1.5) -- (11.3, 1, 1);
    
    \DrawBox{1.1}{0}{2}{.4}{1}{1}{c1}
    \DrawBox{1.6}{0}{2}{.4}{1}{1}{c2}
    \draw[->, ultra thick, draw=red] (2, .6, 2.5) -- (2.5, .5, 3);
    \draw[->, ultra thick, draw=black] (2, .6, 2.5) -- (8.6, .6, 2.5);
    \node[anchor=west] at (2.1, .95, 2.5) {\tiny $H/4 \times W/4 \times S \times 4C$};
    
    \DrawBox{8.6}{0}{2}{.4}{1}{1}{c2}
    \DrawBox{9}{0}{2}{.8}{1}{1}{c1}
    \DrawBox{9.9}{0}{2}{.4}{1}{1}{c2}
    \draw[->, ultra thick, draw=blue] (10.3, .6, 2.5) -- (10.6, .5, 2);
    
    \DrawBox{2.1}{0}{3}{.8}{.5}{1}{c3}
    \DrawBox{3}{0}{3}{.8}{.5}{1}{c4}
    \draw[->, ultra thick, draw=red] (3.8, .3, 3.5) -- (4.3, .25, 4);
    \draw[->, ultra thick, draw=black] (3.8, .3, 3.5) -- (5.8, .3, 3.5);
    \node[anchor=east] at (2.5, -.5, 3.1) {\tiny $H/8 \times W/8 \times S \times 8C$};
    
    \DrawBox{5.8}{0}{3}{.8}{.5}{1}{c4}
    \DrawBox{6.6}{0}{3}{1.6}{.5}{1}{c3}
    \DrawBox{8.3}{0}{3}{.8}{.5}{1}{c4}
    \draw[->, ultra thick, draw=blue] (9.1, .3, 3.5) -- (9.4, .25, 3);
    
    \DrawBox{3.5}{0}{4}{1.6}{.25}{1}{c1}
    \DrawBox{5.2}{0}{4}{1.6}{.25}{1}{c2}
    \draw[->, ultra thick, draw=blue] (6.8, .13, 4.5) -- (7.4, .125, 4);
    \node at (5.1, -.5, 4.1) {\tiny $H/16 \times W/16 \times S \times 16C$};
    
    \draw[->, ultra thick, draw=red]  (11, .2, 4) -- node[above] {\scriptsize subsample} (12.5, .2, 4);
    \draw[->, ultra thick, draw=blue]  (11, -.5, 4) -- node[above] {\scriptsize upsample} (12.5, -.5, 4);
    \end{tikzpicture}

\caption{Illustration of the SEU-Net architecture. We parametrize it by the height (i.e. number of subsamplings/upsamplings set to four in this example) and number of filters $C$ in the first layer (after pooling we double the number of filters and after subsampling we halve it). The values $H, W$ represent the height, width of the inputs images, $S$ is the number of scales in the lifting and $C_\text{in}$, are the number of channels in the input image or convolutional filters and $C_\text{out}$ is the number of classes. Two convolutions are performed between subsamplings and between upsamplings. Feature maps connected by skip connections have the same spatial dimensions.
}
\label{fig:unet_illus}
\end{figure}

Recall that the U-Net \cite{ronneberger2015unet}, illustrated in Figure \ref{fig:unet_illus}, is a CNN architecture for semantic segmentation based on an auto-encoder structure with skip connections linking the encoder and decoder.
As such, it has four main components: convolution blocks, pooling, upsampling and skip connections.
In this section we aim to propose the Scale-Equivariant U-Net (SEU-Net), in order to have a U-Net with increased generalization capacity.

In the framework of the previous section, a network can be written as $\Gamma = \Pi\circ\Sigma\circ\Lambda$, where $\Lambda$ and $\Pi$ are the lifting and projection respectively, and $\Sigma$ is the core part of the network mapping the lifted space to itself. We already saw that $\Lambda \circ R^\prime_{\gamma^k, z} = R_{\gamma^k, z} \circ \Lambda$ and we assume $R^\prime_{\gamma^k, z}\circ \Pi \approx \Pi\circ R_{\gamma^k, z}$. Hence, to build a (approximately) scale-equivariant network, it is sufficient to have $\Sigma\circ R_{\gamma^k, z} = R_{\gamma^k, z} \circ \Sigma$.
In particular, a way to render the U-Net scale-equivariant is to design scale-equivariant versions of its components in $\Sigma$. Convolutions are already rendered equivariant by scale-cross-correlations, and pointwise-non-linearities, batch-normalization and skip connections are equivariant as is.
The rest of this section is dedicated to examining the remaining components: subsampling and upsampling.

\subsection{Subsampling}
Classical pooling operators naively applied scale by scale do not result in scale-equivariant poolings in the lifting space. For example, the max-pooling $\text{MP}[f](x) = \max_{y \in N} f(rx + y)$ with strides $r\in\mathbb{N}$ and neighborhood $N \subseteq \Z^2$ (usually a $r \times r$ square). Its naive extension to the lifted space $\text{MP}'[f](\gamma^k, x) = \text{MP}[f(\gamma^k, \cdot)](x) \ \forall k\in\mathbb{N}$ does not commute with $R_{\gamma^k, x}$.

Strided convolutions however, generalize well to this scenario, written as the subsampling operator $D_t[f](\gamma^k,x) = f(\gamma^k, tx)$ following a scale-cross-correlation.
We can verify that it is scale-equivariant: $D_t[R_{\gamma^k, x} f](\gamma^l, y) = (R_{\gamma^k, x} f)(\gamma^l, ty) = f(\gamma^{l+k}, \gamma^kty) = D_t[f](\gamma^{l+k}, \gamma^ky) = R_{\gamma^k, x} [D_t f](\gamma^l, y)$. We use strides as the subsampling in our networks, with a stride of $t = 2$.

\subsection{Upsampling}
\label{sec:upsampling}

Upsampling blocks are a well established part of modern neural network architectures for segmentation and other tasks. In order to extend upsampling to a scale-equivariant setting, we look at the case where $f$ is defined on a continuous domain. In that case, the downsampling $D_{\gamma^l}$ has an inverse $U_{\gamma^l}$ which is the natural upsampling.

In the discrete case the problem becomes more complicated as downscaling is not invertible, but for $k, l\in \mathbb{N}$ we can define an upsampling $U_{\gamma^l}$ as an operator satisfying $\forall x \in \Z^2$
\begin{equation}
    U_{\gamma^k}[f](\gamma^l,\gamma^k x) = f(\gamma^l,x) \quad \text{and} \quad U_{\gamma^{lk}} = U_{\gamma^k} \circ U_{\gamma^l}.
    \label{eq:discrete-upsampling}
\end{equation}
With this, we have $D_{\gamma^k} \circ U_{\gamma^k} = \mathrm{id}$.
For all $k$, $U_{\gamma^k}(f)$ values are only restricted in the points $y \in k \Z^2 = \{kx | x \in \Z^2\}$, and the values on the other pixels can be defined in several ways (\emph{e.g.} copies, interpolation) as long as it satisfies \eqref{eq:discrete-upsampling}.
Now, if $U_{\gamma^l} R_{\gamma^k, x} f = R_{\gamma^k, \gamma^l x} U_{\gamma^l} f$ for any $f$ then $\Sigma\circ R_{\gamma^k, x} = R_{\gamma^k, x}\circ\Sigma$.
Indeed let $\psi_i =  L_i \cdots D_{\gamma^l} L_1 $, $i=1,\dots,m$ denote the part of a SEU-Net of height $m$ before the $i$-th downsampling block, where $L_j$, $j = 1, \dots, m$, are blocks that commute with $R_{\gamma^k, x}$(constructed by scale-cross-correlations, pointwise activations and batch normalization). Denote $\phi_m = L_{m+1} \psi_m$ and $\phi_i = L_{i} C(U_{\gamma^l} \phi_{i+1}, \psi_i)$, $i=m,\dots,1$ where $C$ denotes concatenation. 
With the above hypothesis, we have $\phi_i R_{\gamma^k, x} f = R_{\gamma^k, \gamma^{li} x} \phi_i f$. 
In particular, $R_{\gamma^k, x} \phi_0 f = \phi_0 R_{\gamma^k, x} f$, and we notice that $\phi_0$ is precisely $\Sigma$.

The sufficient condition $U_{\gamma^l} R_{\gamma^k, x} f = R_{\gamma^k, \gamma^l x} U_{\gamma^l} f$ is not verified in general (see Appendix B), but Proposition~\ref{prop:upsamp} introduces a setting where it does. 

\begin{proposition}\label{prop:upsamp}
For $N\in\mathbb{N}^*$ and $i \in\{1, \dots, N\}$, let $\mathcal{U}_i = \{U_{\gamma^{n_i+l}} f_i | l \in \mathbb{N} \}$, where each $f_i:G \to \R^n$ is a function on $G$ and each $n_i$ an integer. Let $n_0 \leq \min \{ n_i | i = 1, \dots, N\}$ and $\mathcal{U} = \bigcup\limits_{i=1}^N \mathcal{U}_i$. 
Then for all $f \in \mathcal{U}$, and $k,l \in \mathbb{N}$ such that $k - l \leq n_0$, we have\footnote{For a proof, see Appendix C.} $U_{\gamma^l} R_{\gamma^k,x} f =  R_{\gamma^k,\gamma^l x} U_{\gamma^l} f$.
\end{proposition}

This property states that upsampling behaves as an equivariant operator as long as the input image is an upsampling of some image in a base scale.
It can be interpreted as saying that the downscaling should not destroy information of the images in $\mathcal{U}$. 
We model this by constraining the scaling factors of the downscaling actions and assuming that the objects of interest in an image are sufficiently big.
We would like to point out that this hypothesis is never verified but reasonable for most of the datasets for semantic segmentation.

Before moving on to the experimental part, let us sum up the theoretical properties of a SEU-Net $\Gamma = \Pi\circ\Sigma\circ\Lambda$. By our construction we can hope for an approximated scale-equivariance $\Gamma\circ R^{\prime}_{\gamma^k,z} \approx  R^{\prime}_{\gamma^k,z} \circ \Gamma$. Two approximations prevent from exact equivariance: The approximated equivariance of the projection operator $\Pi$, which is intrinsic to the lifting approach, and the assumption to guarantee an equivariant upsampling, which is never verified in practice. We will see in our experiments that the SEU-Net shows a high degree of scale-equivariance despite these approximations.
Each of these approximations is intrinsic to the problem. If the problem was formulated in a continuous domain $\mathcal{S}_\gamma \times \R^2$  upsampling would be theoretically equivariant, but its implementation would have the same problems.

\section{Experiments}
\label{sec:experiments}

In this section we test the proposed SEU-Net in two segmentation tasks where we evaluate its generalization to unseen scales. We train the SEU-net on a set where objects have roughly the same scale and test it on a wide range of scales.
For these experiments we use a scale base of $\gamma = 2$, downsampling $D_2$ and upsampling $U_2$ computed by bilinear interpolation.
Quantitative results will be measured using Intersection over Union (IoU) and consistency. We define consistency as follows: given a segmentation neural network  $\phi$, the consistency is the probability of assigning the same label to a pixel after it has been transformed, formally $\text{Cons}(\phi, s) = P\Big(\phi(R_{s,0}[f])(x) = R_{s,0}[\phi(f)](x)\Big)$. 

We compare the SEU-Net to U-Net and to the SResNet \cite{worrall2019scale}, a scale-equivariant residual architecture which applies subsampling but no upsampling inside the equivariant pipeline, i.e. it only applies an upsampling \emph{after} the projection layer. Hence, it does not benefit from the features that made U-Net more suitable for segmentation, namely the skip connections at several upsampling stages.

\textbf{Scale Dropout. }
In order to produce more robust results with respect to scale changes we propose the use of \emph{Scale Dropout} before the projection layers. Given a feature map $f:S_\gamma \times \Z^2 \to \R^n$, we compute its scale dropout of rate $p\in [0,\ 1]$ as $\text{ScaleDropout}_p(f)(s, x) = X(s) f(s, x)$ where $X(s)$ is a Bernoulli variable of parameter $p$, i.e. $P\big(X(s) = 1\big) = 1-p$ and $P\big(X(s) = 0\big) = p$. In our experiments we use values $p = 0$ (no dropout) and $p = 0.25$.

\subsection{Oxford-IIIT Pet Dataset}
\label{sec:oxford_pets}
\begin{figure}[th]
    \centering
    \begin{minipage}[t]{.22\textwidth}
    \centering
    \includegraphics[width=.125\textwidth]{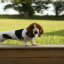}
    \includegraphics[width=.25\textwidth]{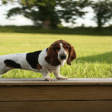}
    \includegraphics[width=.5\textwidth]{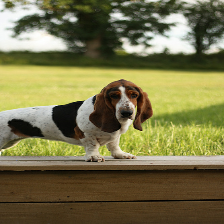}
    
    \includegraphics[width=1.\textwidth]{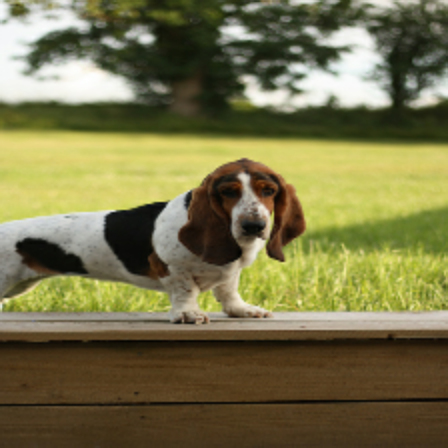}
    
    (a) Image
    \end{minipage}
    \begin{minipage}[t]{.22\textwidth}
    \centering
    \includegraphics[width=.125\textwidth]{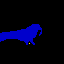}
    \includegraphics[width=.25\textwidth]{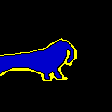}
    \includegraphics[width=.5\textwidth]{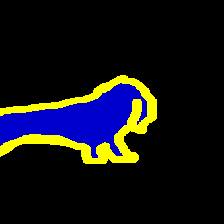}
    
    \includegraphics[width=1.\textwidth]{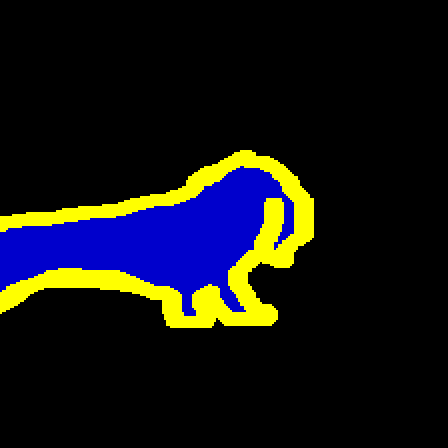}
    
    (b) Ground Truth
    \end{minipage}
    ~
    \begin{minipage}[t]{.22\textwidth}
    \centering
    \includegraphics[width=.125\textwidth]{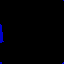}
    \includegraphics[width=.25\textwidth]{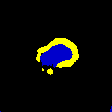}
    \includegraphics[width=.5\textwidth]{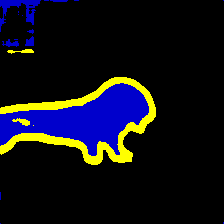}
    
    \includegraphics[width=1.\textwidth]{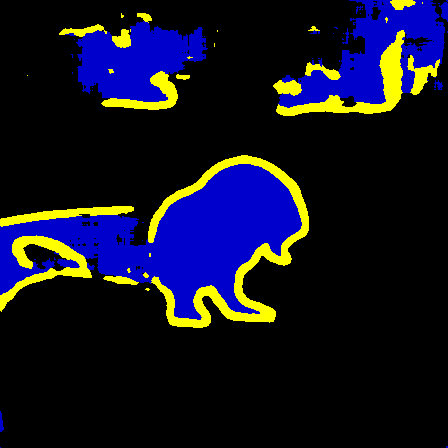}
    
    (c) U-Net
    \end{minipage}
    \begin{minipage}[t]{.22\textwidth}
    \centering
    \includegraphics[width=.125\textwidth]{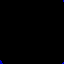}
    \includegraphics[width=.25\textwidth]{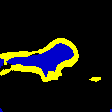}
    \includegraphics[width=.5\textwidth]{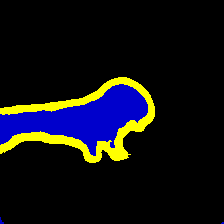}
    
    \includegraphics[width=1.\textwidth]{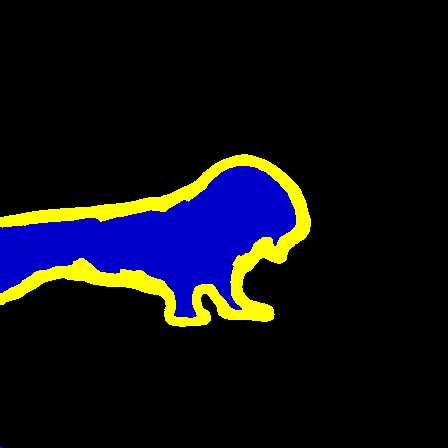}
    
    (d) SEU-Net
    \end{minipage}
    \caption{Sample test image at different scales and ground truth from the Oxford-IIIT Pet dataset, along with the U-Net and SEU-Net predictions. The scales present are $0.25, 0.5, 1$ and $2$ times the training scale.}
    \label{fig:example_pets}
\end{figure}

The Oxford-IIIT Pet~\footnote{\url{https://www.robots.ox.ac.uk/~vgg/data/pets/}, CC BY-SA 4.0 license} dataset \cite{parkhi12a} consists of pictures containing cats and dogs. The relevant labeling for this paper, the trimaps, is the segmentation of the images into three classes: the animal, the background and the boundaries of the animal.
In Figure \ref{fig:example_pets}(a) and (b) we see an example of an image and its corresponding ground truth.
The dataset was loaded from the TensorFlow package\footnote{\url{https://www.tensorflow.org/datasets/catalog/oxford_iiit_pet}}, where it is divided into $3680$ training samples and $3669$ test samples. To make the validation set we removed $200$ test samples.
During training and testing images are resized to $224 \times 224$ pixels.
We define multiple test sets by re-scaling the original test set by $s\in \{2^{\frac{i}{2}} | i \in \{-4,-3,\dots,4\} \}$. We used bilinear interpolation to up-scale images.

Both the U-Net and SEU-Net have height four and contain sixteen filters in the first layer and use $3 \times 3$ filters. The SEU-Net truncates at four scales, and filters have depth one in the scales dimension (their values is different from zero in one scale value). The networks are trained using the Adam \cite{kingma2014adam} optimizer with categorical cross-entropy loss. Training the U-Net, SResNet and SEU-Net takes approximately $24$, $73$, and $97$ seconds per epoch respectively, on a Tesla P100-SXM2-16Gb GPU.

\textbf{Comparison with data augmentation.} We also performed scale jittering in the U-Net to compare the effect of the equivariant network with the effect of data augmentation. Scale jittering is performed by rescaling the image by a randomly chosen scale $\alpha$ and either random cropping or padding to the original image. We trained a U-Net with scale jittering in the interval $[\frac{1}{4}, 4]$, equal to the interval of test scales.

\begin{figure}[t]
    \centering
    \begin{minipage}{.49 \linewidth}
    \centering
    \includegraphics[width=\linewidth]{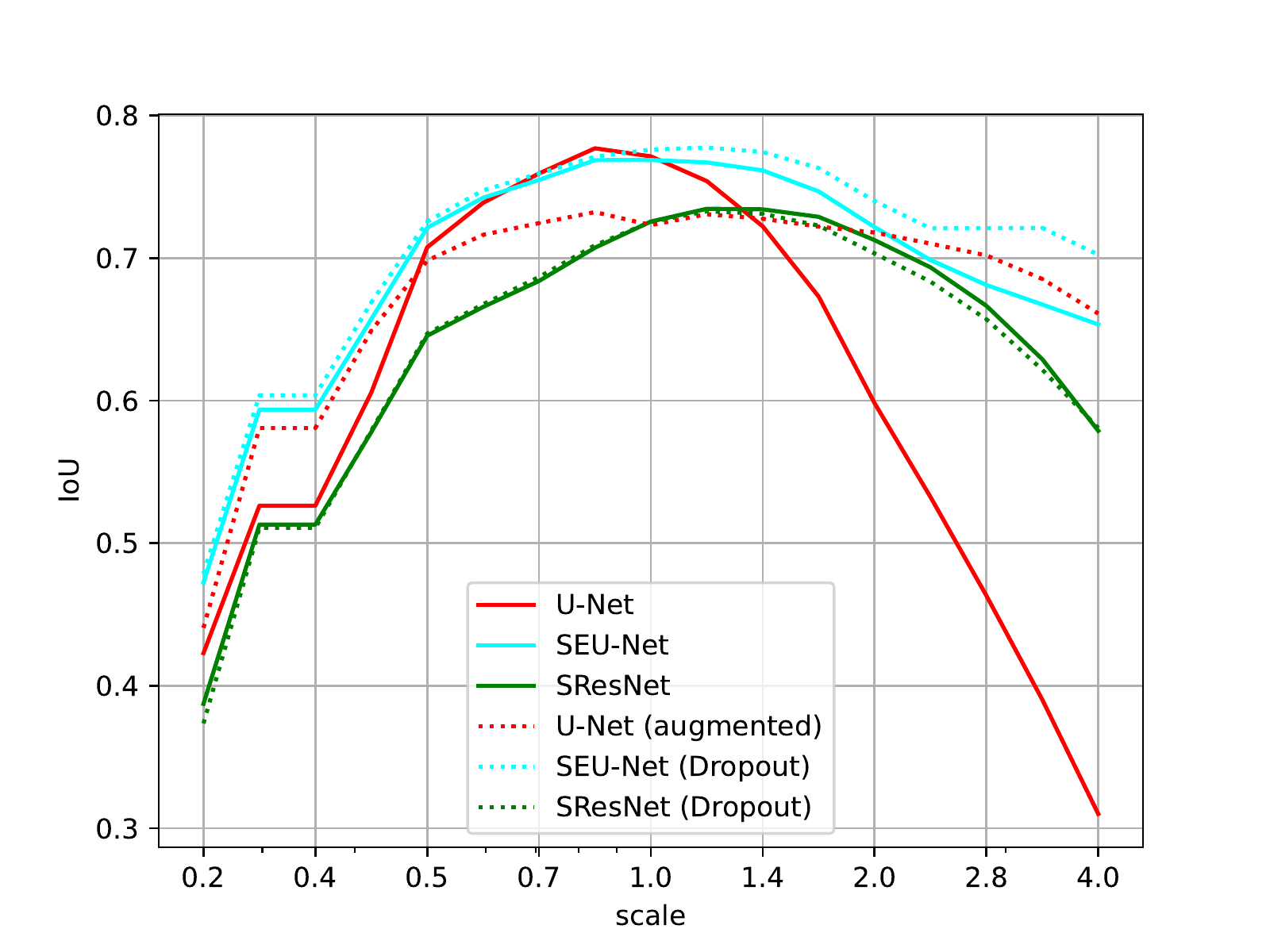}
    
    (a) IoU
    \end{minipage}
    \begin{minipage}{.49 \linewidth}
    \centering
    \includegraphics[width=\linewidth]{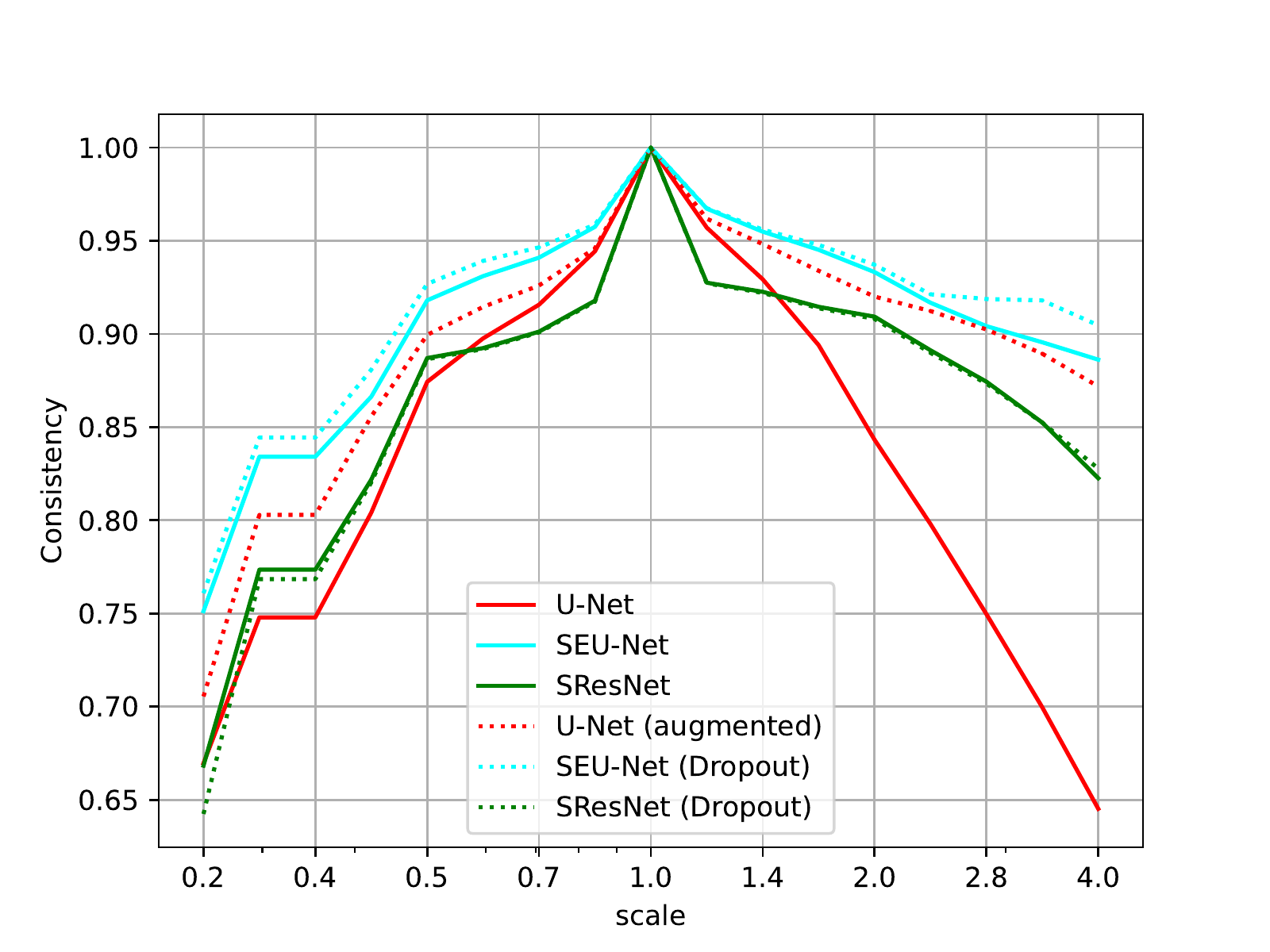}
    
    (b) Consistency
    \end{minipage}
    \caption{Overall results in terms of IoU and Consistency for each scale of the Pet dataset.}
    \label{fig:pet_iou_cons}
\end{figure}

\textbf{Results.} The overall results in terms of the IoU are shown in Figure \ref{fig:pet_iou_cons}. Firstly we notice that the SEU-Net increases performance compared to both SResNet and U-Net. SResNet, however, does not consistently generalize better than the U-Net. Dropout improves the quality of SEU-Net, particularly for more extreme scales, indeed, for larger scales the augmented U-Net has a better IoU than the SEU-Net without scale dropout, but not than the one with scale dropout. The augmented U-Net loses performance scale $1$, it would probably need to be larger to retain the same performance.
We show examples of the predictions of the U-Net and SEU-Net in Figure \ref{fig:example_pets}.

\subsection{Cell Segmentation}
\begin{figure}[th]
    \centering
    \begin{minipage}[t]{.22\textwidth}
    \centering
    \includegraphics[width=.2\textwidth]{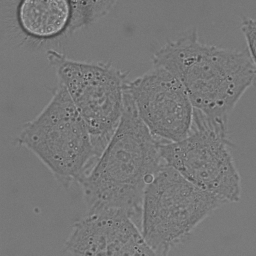}
    \includegraphics[width=.4\textwidth]{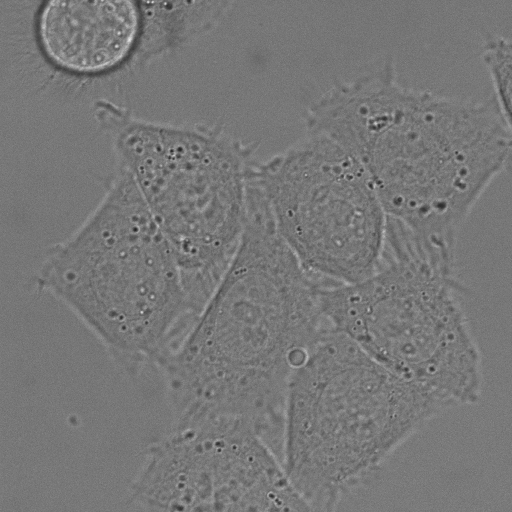}
    
    \includegraphics[width=.8\textwidth]{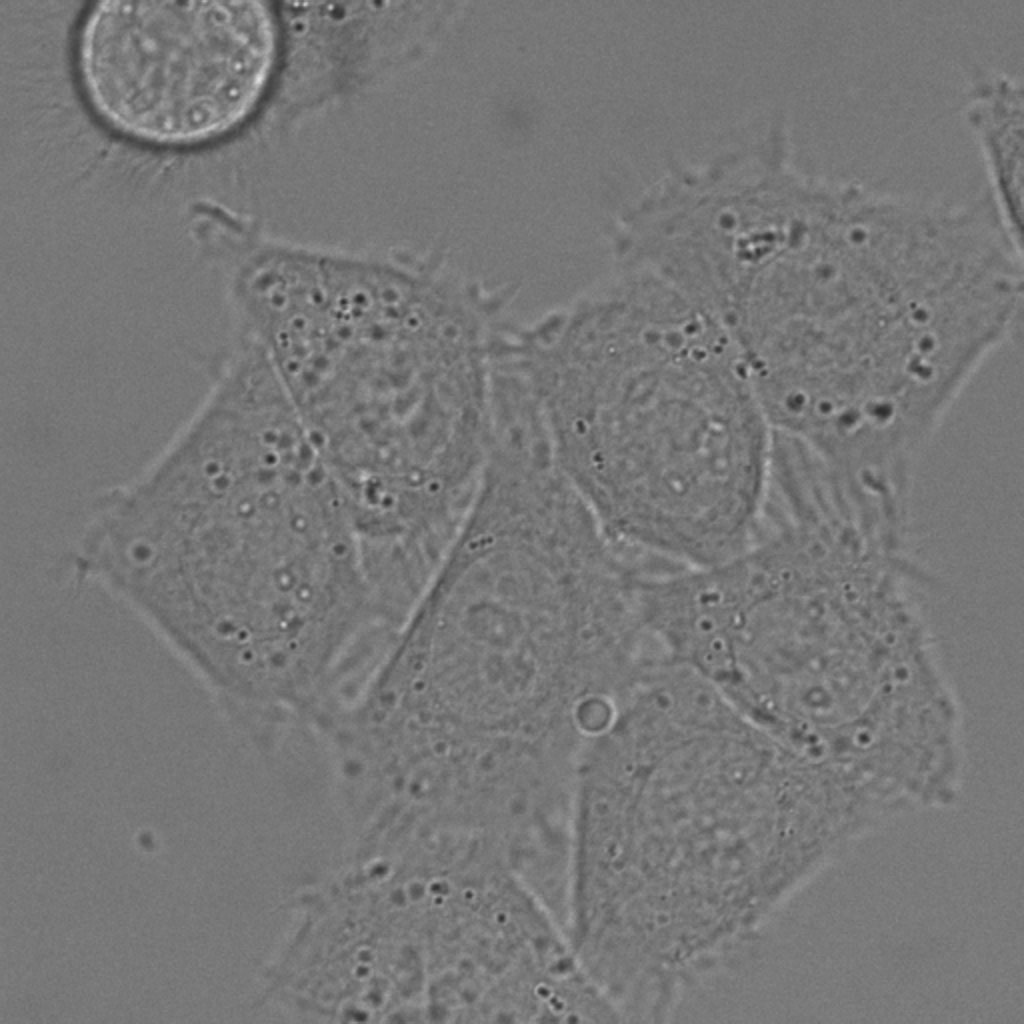}
    
    (a) Image
    \end{minipage}
    \begin{minipage}[t]{.22\textwidth}
    \centering
    \includegraphics[width=.2\textwidth]{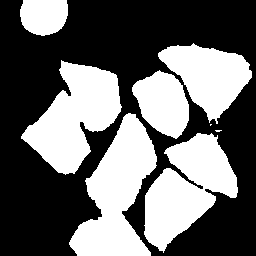}
    \includegraphics[width=.4\textwidth]{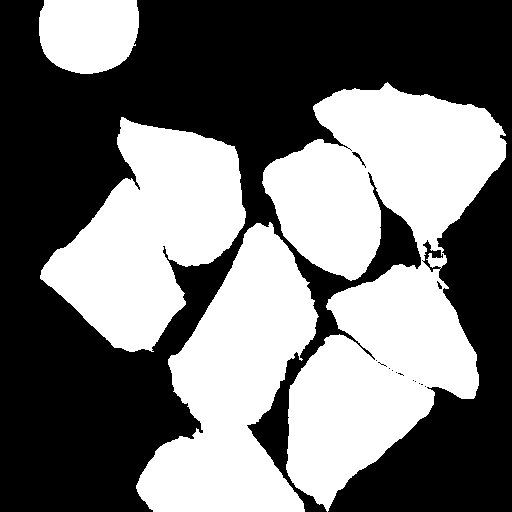}
    
    \includegraphics[width=.8\textwidth]{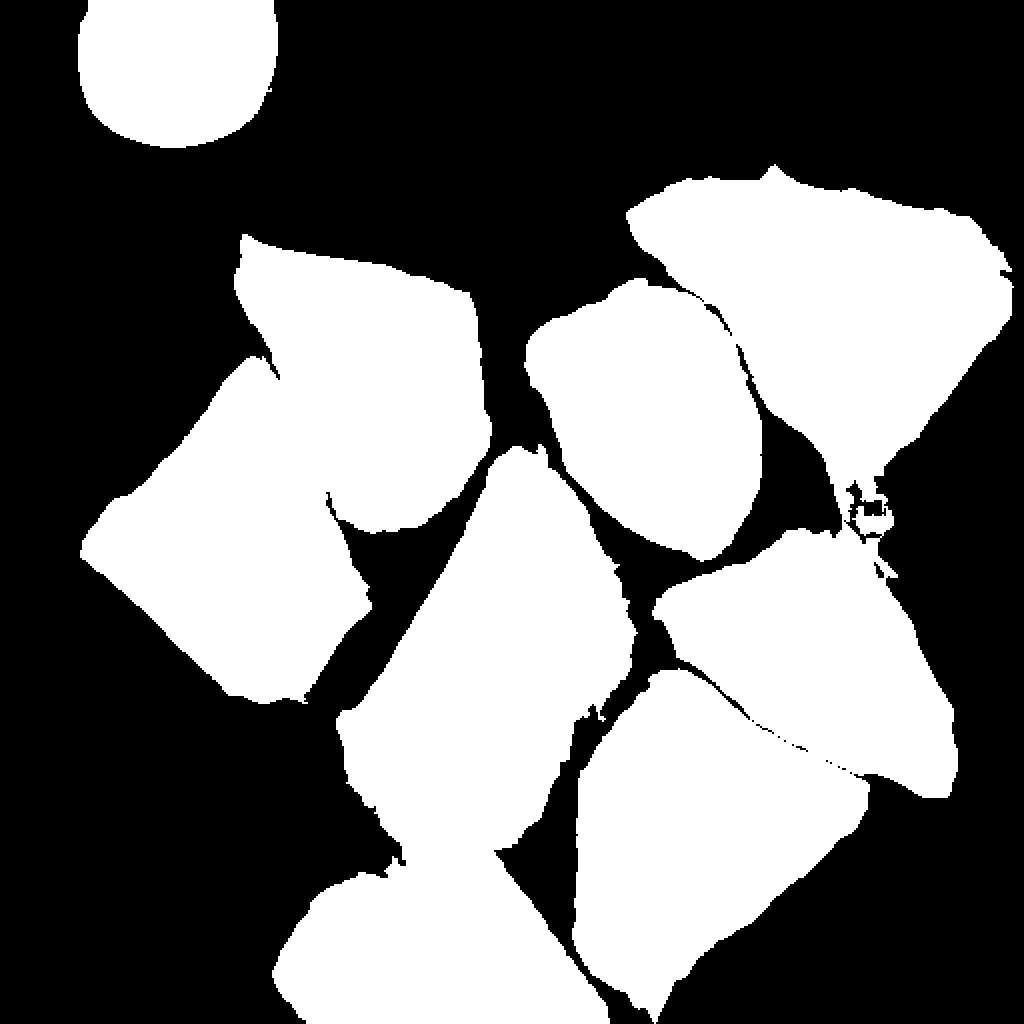}
    
    (b) Ground Truth
    \end{minipage}
    ~
    \begin{minipage}[t]{.22\textwidth}
    \centering
    \includegraphics[width=.2\textwidth]{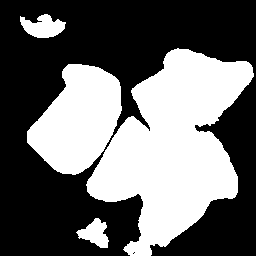}
    \includegraphics[width=.4\textwidth]{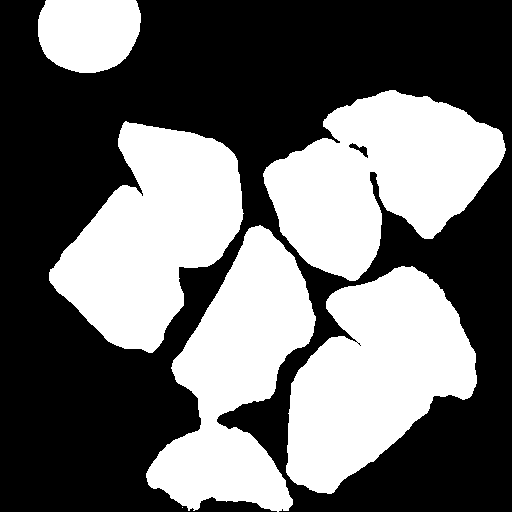}
    
    \includegraphics[width=.8\textwidth]{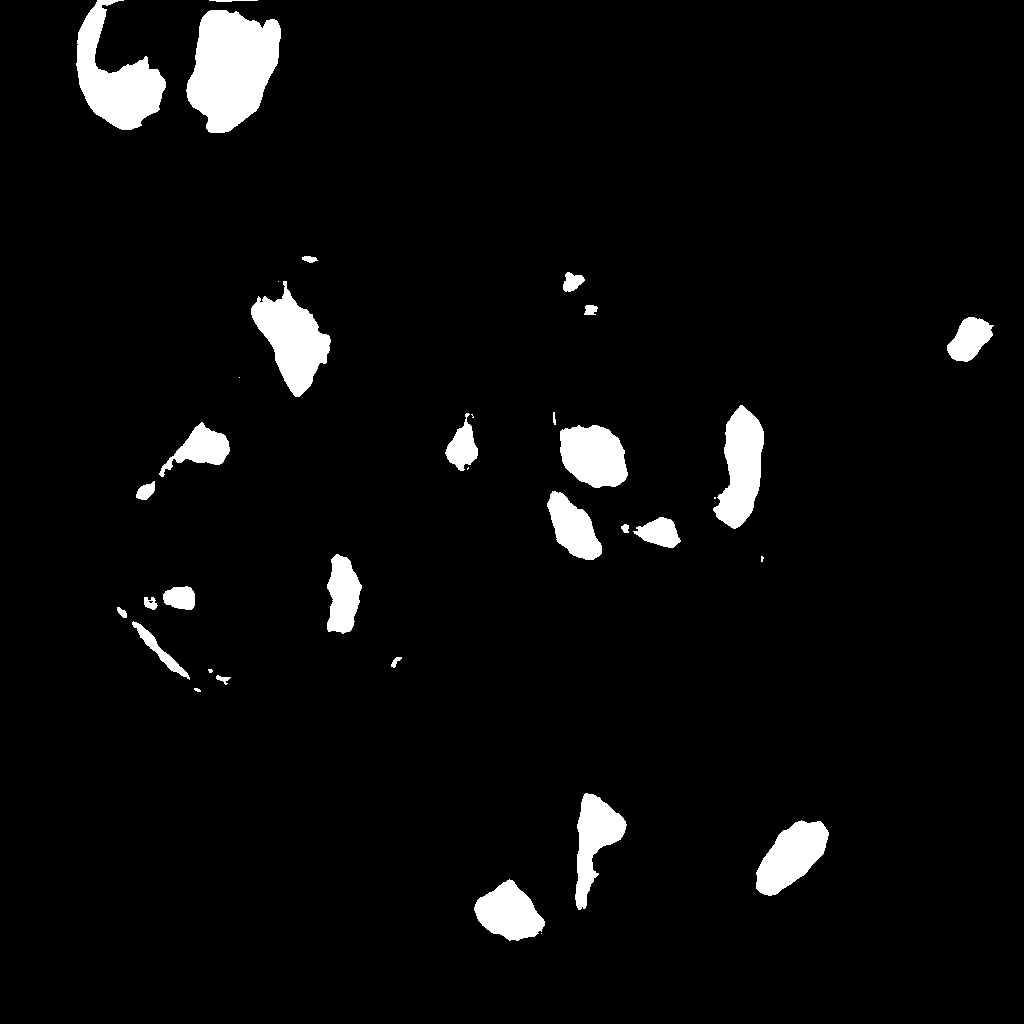}
    
    (c) U-Net
    \end{minipage}
    \begin{minipage}[t]{.22\textwidth}
    \centering
    \includegraphics[width=.2\textwidth]{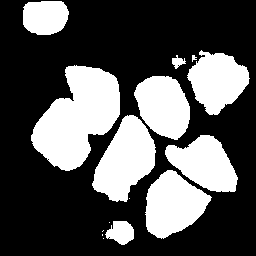}
    \includegraphics[width=.4\textwidth]{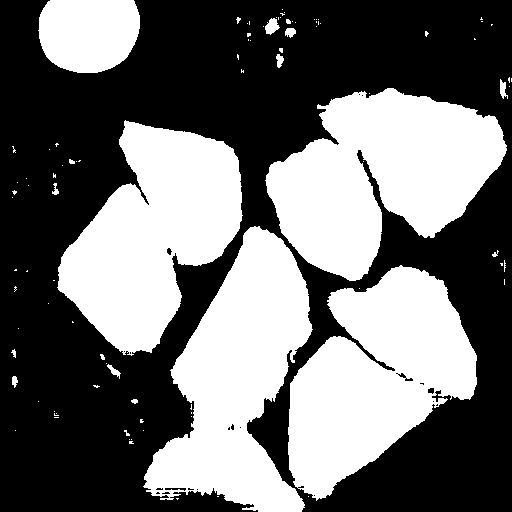}
    
    \includegraphics[width=.8\textwidth]{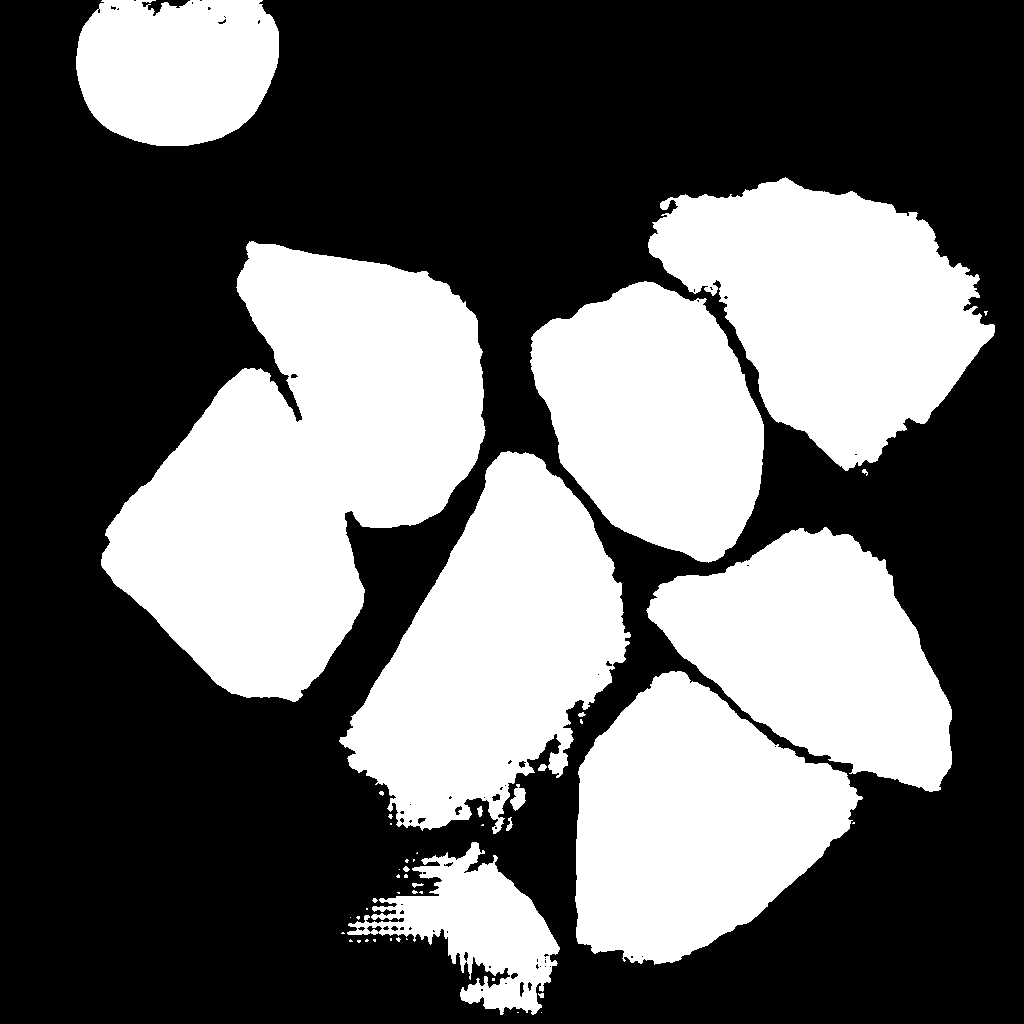}
    
    (d) SEU-Net
    \end{minipage}
    \caption{Predictions from DIC-HeLa at different scales, namely scales $0.5$, $1$ and $2$. Although the U-Net suffers from the scale change, especially the change to a larger scale, the SEU-Net can still capture the overall structure of the cells.}
    \label{fig:example_cells}
\end{figure}

We also evaluate the models in a medical image segmentation dataset, namely the DIC-C2DH-HeLa dataset \cite{ulman2017objective} of HeLa cells on a flat glass recorded by differential interference contrast (DIC). We used $83$ images for train/validation and $83$ for testing. Figure~\ref{fig:example_cells} (a) and (b) shows an example from the test set with its labels at different scales.
\begin{figure}[th]
    \centering
    \begin{minipage}{.49\textwidth}
        \centering
        \includegraphics[width=.99\textwidth]{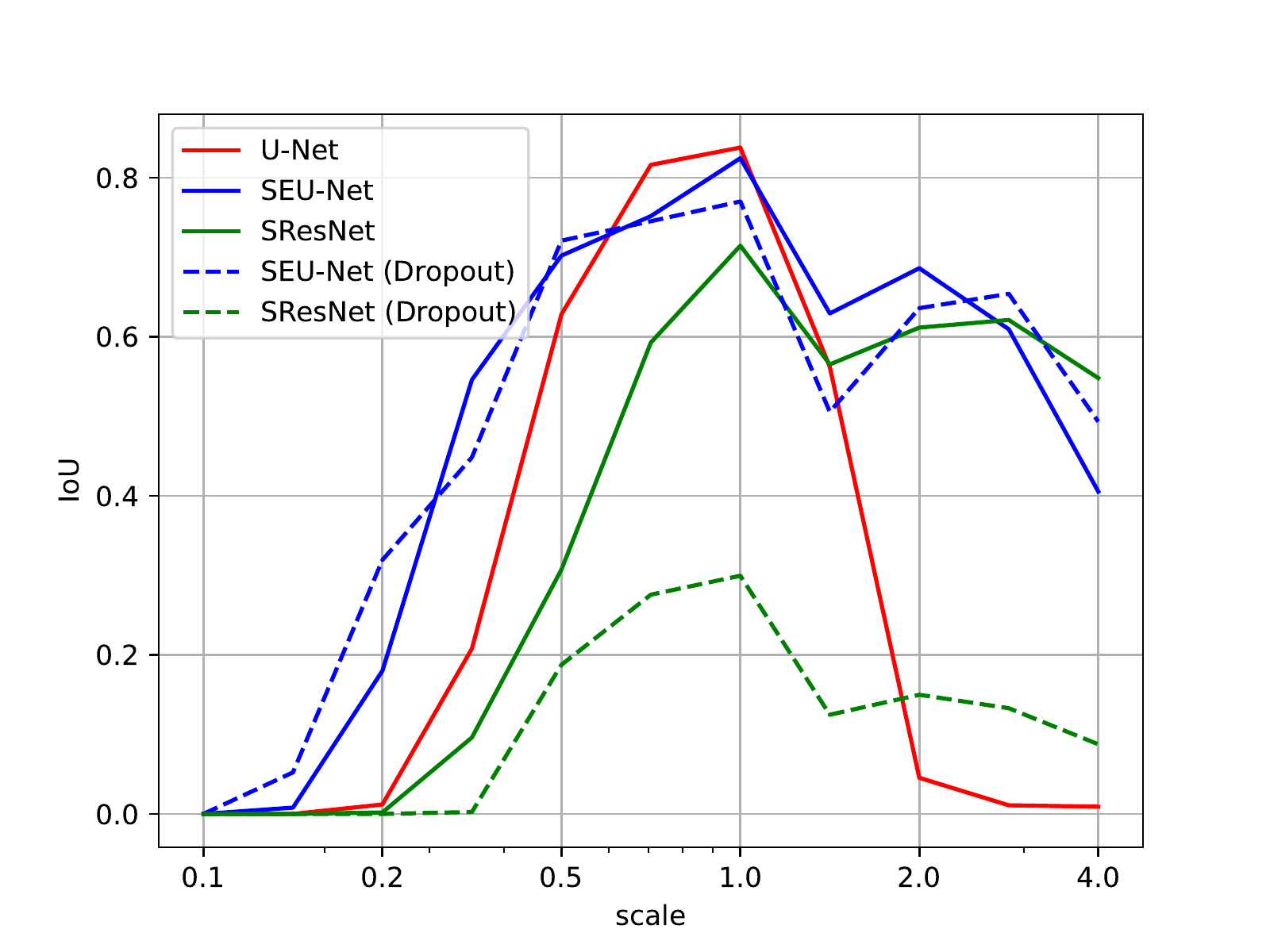}
        
        (a) IoU per scale.
    \end{minipage}
    \begin{minipage}{.49\textwidth}
        \centering
        \includegraphics[width=.99\textwidth]{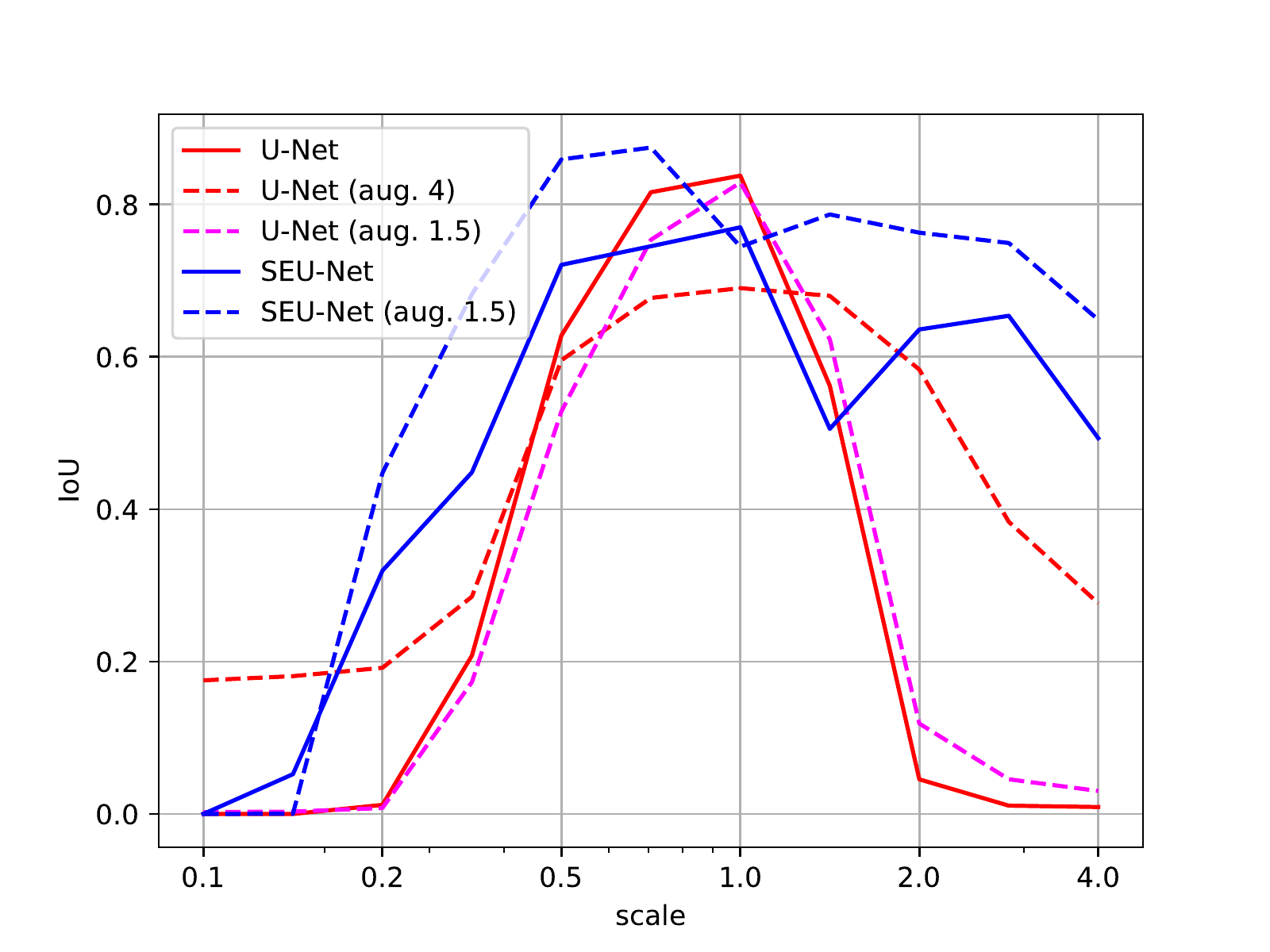}
        
        (b) Augmentation comparison.
    \end{minipage}
    \caption{IoUs of the cell segmentation experiment with comparisons with U-Net, SResNet and data augmentation. U-Net (aug. 4) refers to the U-Net trained with scale jittering with range $4$ and U-Net (aug. $1.5$) refers to the U-Net trained with jittering with range $1.5$. The same for SEU-Net (aug. $1.5$).}
    \label{fig:iou_cell}
\end{figure}
Models are trained with the AdamW optimizer \cite{loshchilov2018decoupled}.
Like in the previous experiment, we first train the models in the training set and test in the test set re-scaled by different values.
We also perform scale jittering, but now for both U-Net and SEU-Net. For U-Net we trained models with scale jittering with ranges $4$ ($\alpha$ is chosen each step from the interval $[\frac{1}{4}, 4]$) and $1.5$ ($\alpha$ is chosen from the interval $[\frac{2}{3}, \frac{3}{2}]$) and for SEU-Net we used only the range $1.5$ jittering.

\textbf{Results.} Figure~\ref{fig:iou_cell} (a) shows the IoU of different models on the re-scaled test sets. Figure~\ref{fig:example_cells} shows some segmentation examples. Again, the SEU-Net outperforms the U-Net. The poor results of the SResNet for smaller scales is possibly due to the cell images containing more high-frequency information, compared to the pets images.
In contrast to the previous experiment, dropout did not seem to significantly increase performance of the SEU-Net, neither in the train scale nor the test scales. Moreover the SResNet results were greatly decreased due to dropout.
This is likely a result of the agumented dataset being more difficult to segment than the original and not being representative of the dataset at base scale.
The gain in generalization is only better than the SEU-Net for the smallest scales. The jittering with range $1.5$ does not have a very noticeable effect.
On the other hand the SEU-Net with $1.5$ jittering has a noticeable gain in generalization to larger scales.

\section{Conclusions and Future Work}
\label{sec:conclusions}
In this paper we revisited the framework of scale semigroup-equivariant neural networks and applied it to the definition of a Scale Equivariant U-Net for semantic segmentation. 
Experimental results show that the SEU-Net can greatly improve the generalization to new scales and even the performance in the training scale. Moreover, the results lead us to conjecture that the U-Net with scale jittering would need to have more parameters to have a good performance in all the range of scales, while the SEU-Net achieves good results without increasing its size. 
The results suggest that implementing that the improvement comes not only from the scale-equivariant cross-correlations, but also from the SEU-Net global architecture and applying the pooling operators inside the equivariant pipeline. The proposed scale dropout was also shown to have the potential to increase scale-equivariant models' performance. In future works it would be interesting to study an equivariant regularization term such as in \cite{sosnovik2021disco} in addition to the scale-dropout.

\section*{Acknowledgements}
This work was granted access to the HPC resources of IDRIS under the allocations AD l011013637 and \\ AD 011012212R1.

\bibliographystyle{abbrv}
\bibliography{references}

\begin{thebibliography}{10}

\bibitem{chidester2019enhanced}
B.~Chidester, T.-V. Ton, M.-T. Tran, J.~Ma, and M.~N. Do.
\newblock Enhanced rotation-equivariant u-net for nuclear segmentation.
\newblock In {\em Proceedings of the IEEE/CVF Conference on Computer Vision and
  Pattern Recognition (CVPR) Workshops}, June 2019.

\bibitem{cohen2016group}
T.~Cohen and M.~Welling.
\newblock Group equivariant convolutional networks.
\newblock In {\em International conference on machine learning}, pages
  2990--2999, 2016.

\bibitem{ghosh2019scale}
R.~Ghosh and A.~K. Gupta.
\newblock Scale steerable filters for locally scale-invariant convolutional
  neural networks.
\newblock {\em arXiv preprint arXiv:1906.03861}, 2019.

\bibitem{heijmans02scale}
H.~J. Heijmans and R.~van~den Boomgaard.
\newblock Algebraic framework for linear and morphological scale-spaces.
\newblock {\em Journal of Visual Communication and Image Representation},
  13(1-2):269--301, 2002.

\bibitem{jansson2020unseenscales}
Y.~Jansson and T.~Lindeberg.
\newblock Exploring the ability of cnns to generalise to previously unseen
  scales over wide scale ranges.
\newblock {\em arXiv preprint arXiv:2004.01536}, 2020.

\bibitem{kingma2014adam}
D.~P. Kingma and J.~Ba.
\newblock Adam: A method for stochastic optimization.
\newblock {\em arXiv preprint arXiv:1412.6980}, 2014.

\bibitem{lindeberg2021scale}
T.~Lindeberg.
\newblock Scale-covariant and scale-invariant gaussian derivative networks.
\newblock In {\em International Conference on Scale Space and Variational
  Methods in Computer Vision}, pages 3--14. Springer, 2021.

\bibitem{loshchilov2018decoupled}
I.~Loshchilov and F.~Hutter.
\newblock Decoupled weight decay regularization.
\newblock In {\em International Conference on Learning Representations}, 2018.

\bibitem{parkhi12a}
O.~M. Parkhi, A.~Vedaldi, A.~Zisserman, and C.~V. Jawahar.
\newblock Cats and dogs.
\newblock In {\em IEEE Conference on Computer Vision and Pattern Recognition},
  2012.

\bibitem{ronneberger2015unet}
O.~Ronneberger, P.~Fischer, and T.~Brox.
\newblock U-net: Convolutional networks for biomedical image segmentation.
\newblock In {\em International Conference on Medical image computing and
  computer-assisted intervention}, pages 234--241. Springer, 2015.

\bibitem{sangalli2021scale}
M.~Sangalli, S.~Blusseau, S.~Velasco-Forero, and J.~Angulo.
\newblock Scale equivariant neural networks with morphological scale-spaces.
\newblock In {\em International Conference on Discrete Geometry and
  Mathematical Morphology}, pages 483--495. Springer, 2021.

\bibitem{sosnovik2021disco}
I.~Sosnovik, A.~Moskalev, and A.~Smeulders.
\newblock Disco: accurate discrete scale convolutions.
\newblock In {\em Proceedings of the 32nd British Machine Vision Conference},
  2021.

\bibitem{sosnovik2019scale}
I.~Sosnovik, M.~Szmaja, and A.~Smeulders.
\newblock Scale-equivariant steerable networks.
\newblock In {\em International Conference on Learning Representations}, 2019.

\bibitem{thomas2018tensor}
N.~Thomas, T.~Smidt, S.~Kearnes, L.~Yang, L.~Li, K.~Kohlhoff, and P.~Riley.
\newblock Tensor field networks: Rotation-and translation-equivariant neural
  networks for 3d point clouds.
\newblock {\em arXiv preprint arXiv:1802.08219}, 2018.

\bibitem{ulman2017objective}
V.~Ulman, M.~Ma{\v{s}}ka, K.~E. Magnusson, O.~Ronneberger, C.~Haubold,
  N.~Harder, P.~Matula, P.~Matula, D.~Svoboda, M.~Radojevic, et~al.
\newblock An objective comparison of cell-tracking algorithms.
\newblock {\em Nature methods}, 14(12):1141--1152, 2017.

\bibitem{boomgaard94}
R.~Van Den~Boomgaard and A.~Smeulders.
\newblock The morphological structure of images: The differential equations of
  morphological scale-space.
\newblock {\em IEEE Transactions on Pattern Analysis and Machine Intelligence},
  16(11):1101--1113, 1994.

\bibitem{weiler20183d}
M.~Weiler, M.~Geiger, M.~Welling, W.~Boomsma, and T.~S. Cohen.
\newblock 3d steerable cnns: Learning rotationally equivariant features in
  volumetric data.
\newblock In {\em Advances in Neural Information Processing Systems}, pages
  10381--10392, 2018.

\bibitem{weiler2018steerable}
M.~Weiler, F.~A. Hamprecht, and M.~Storath.
\newblock Learning steerable filters for rotation equivariant cnns.
\newblock In {\em Proceedings of the IEEE Conference on Computer Vision and
  Pattern Recognition}, pages 849--858, 2018.

\bibitem{worrall2019scale}
D.~Worrall and M.~Welling.
\newblock Deep scale-spaces: Equivariance over scale.
\newblock In {\em Advances in Neural Information Processing Systems}, pages
  7364--7376, 2019.

\bibitem{zhu2019scale}
W.~Zhu, Q.~Qiu, R.~Calderbank, G.~Sapiro, and X.~Cheng.
\newblock Scale-equivariant neural networks with decomposed convolutional
  filters.
\newblock {\em arXiv preprint arXiv:1909.11193}, 2019.

\end{thebibliography}

\appendix

\section{Semigroup cross-correlation}
Let $\mathcal{F} = \mathbb{R}^G$ denote the set of functions mapping
$G$ to $\mathbb{R}$. Bearing in mind the final purpose of defining
equivariant CNN layers, we focus on linear operators on
$\mathcal{F}$. Let the semigroup right action $(R_u)_{u\in G}$ on
$\mathcal{F}$, defined by
\begin{equation}
\label{eq:right-action-semigroup}
\forall u,g\in G, \forall f\in\mathcal{F},\;\;\; R_u(f)(g) = f(u\cdot g).
\end{equation}
Then for any $h\in\mathcal{F}$, the linear operator defined by
\begin{equation}
 \label{eq:semigroup-convolution-new}
 \forall u\in G, \;\;\;  H(f)(u) = (f \star_G h)= \sum_{g\in G} R_u(f)(g) h(g)
\end{equation}
is equivariant to the semigroup action $(R_u)_{u\in G}$, as
$H(R_u(f)) = R_u(H(f)).$ This class of semigroup equivariant linear
operators is the \emph{semigroup cross-correlation} proposed
in~\cite{worrall2019scale} as the key element to define
scale-equivariant convolutional layers. Note that when $G$ is the
group of image translations (groups are special semigroups),
\eqref{eq:semigroup-convolution-new} corresponds to the classic
discrete convolution with the reversed filter $h^*(g) = h(g^{-1})$. We
use the notation $f \star_G h$ remarking however that this operation
is not symmetrical in $f$ and $h$ even when the law $\cdot$ on $G$ is
commutative.  Also, contrary to the group case, we do not have the
property that every linear and equivariant operator can be written as
a semigroup-cross-correlation.

\section{Different Pooling Operators}

Besides the strided scale-crosscorrelations we used, we can define another class of pooling operators, inspired by classical max-pooling.
Let us place ourselves in a slightly different case of pooling a function in a continuous domain $f:\mathcal{S} \times \R^2 \to \R$, with $\mathcal{S} \times \R^2$ acting on it by $R_{\gamma^l, z} f (\gamma^k, x) = f(\gamma^{k+l}, \gamma^l x + z)$, $k,l \in \mathbb{N}$, $x \in \R^2$, $z \in \Z^2$.
We define the pooling of $f$ as an operator $F$ followed by a downsampling $D_{\gamma^l}[f](\gamma^k,z) = f(\gamma^k, \gamma^l z)$
\begin{equation}
    P[f] = D_{\gamma^l} F f.
\end{equation}
If $F$ commutes with $R_{\gamma^k, x}$, then so does $P$. We consider three pooling functions: $F_\mathrm{id}=\mathrm{id}$ (strides) and two dilation scale-spaces \cite{heijmans02scale}:
\begin{itemize}
    \item  The max-pooling of scale-semigroup-valued images is given by a re-scaled max-pooling
\begin{equation}
\label{eq:dilation}
    F_{\text{max}}[f](\gamma^k, z) = \sup\limits_{y \in N_k \times N_k} \{ f(z-y) \}
\end{equation}
where $N_k = \{\gamma^k x | x \in N\}$ and $N$ is for example a $\gamma^l$-sided square in $\R^2$. 

    \item The quadratic dilation (quadpool) scale-space is a morphological counterpart to the Gaussian scale-space \cite{boomgaard94} defined by
\begin{equation}
    \label{eq:quad_dilation}
    F_{\text{quad}}[f](\gamma^k, z) = \sup\limits_{y \in \R^2} \left\{f(z-y) - \frac{\lVert y \rVert^2}{c \gamma^{2k}} \right\},
\end{equation}
where $c > 0$ is some constant.
\end{itemize}

In contrast to the strided scale-cross-correlations given by $F_\mathrm{id}$, the functions $F_{\max}$ and $F_{\mathrm{quad}}$ are scale-equivariant only in this continuous setting, their discretized versions are not actually equivariant. Nonetheless, a network employing scale-cross-correlations and these poolings would be equivariant when applied to signals in the domain $\mathcal{S} \times \R^2$.

\begin{figure}[t]
    \centering
    \begin{minipage}{.49 \linewidth}
    \centering
    \includegraphics[width=\linewidth]{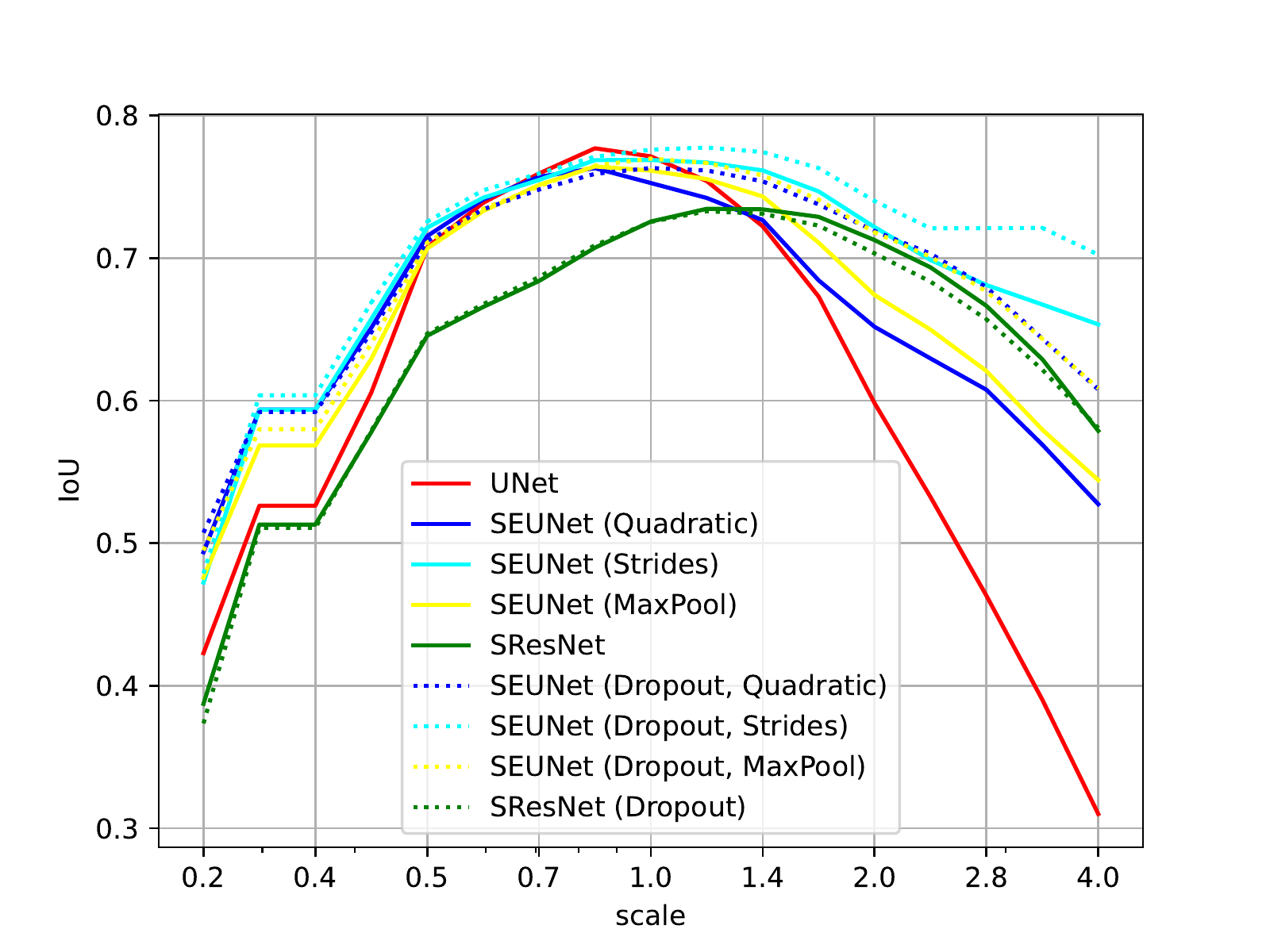}
    
    (a) IoU
    \end{minipage}
    \begin{minipage}{.49 \linewidth}
    \centering
    \includegraphics[width=\linewidth]{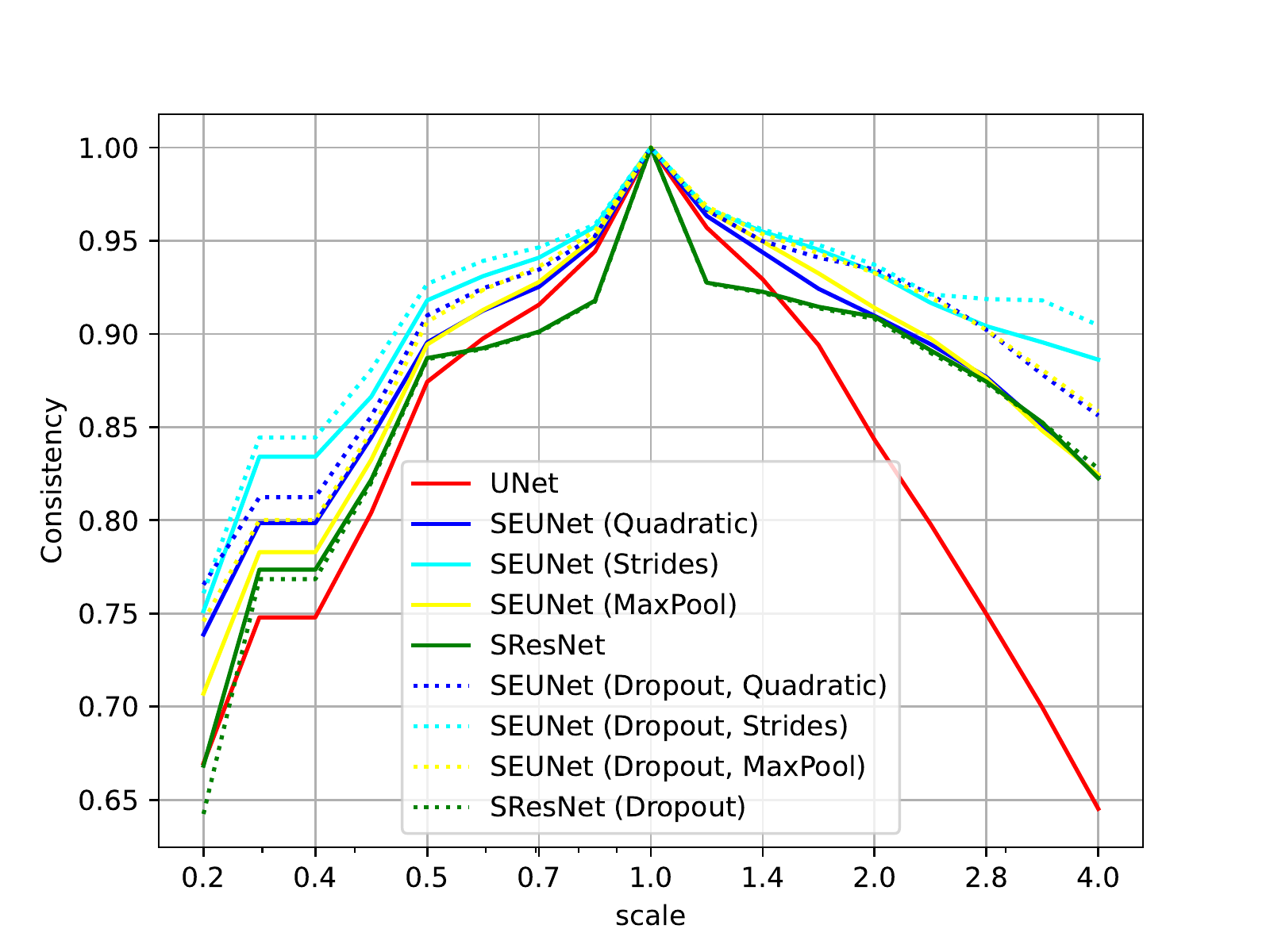}
    
    (b) Consistency
    \end{minipage}
    \caption{Overall results in terms of IoU and Consistency for each scale of the Pet dataset. The SEU-Net has been computed with different pooling functions.}
    \label{fig:pet_iou_cons_poolings}
\end{figure}

In Figure \ref{fig:pet_iou_cons_poolings} we extend the experiments from Section 5.2. Using different pooling functions did not improve the performance of the SEU-Net compared to its performance using strided scale-cross-correlations.

\section{Non-equivariance of the upsampling}
In this section we show that  $U_{\gamma^l} R_{\gamma^k, x} f \neq R_{\gamma^k, \gamma^l x} U_{\gamma^l} f$ for at least one lifted image $f$, one couple of integers $(k,l)$ and a point $x \in \Z^2$. Note that $U_{\gamma^l}$ is an upsampling defined in the associated paper.

\begin{figure}
    \centering
    \begin{tikzpicture}
        \node (im1) at (0,2) {\includegraphics[width=.06\textwidth]{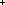}};
        \node (im1_dn) at (3.5,0) {\includegraphics[width=.03\textwidth]{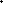}};
        \node (im1_dnup) at (6.5,0) {\includegraphics[width=.06\textwidth]{figures/upsample_ex_im1.png}};
        
        \node (im1_up) at (3.5,2) {\includegraphics[width=.12\textwidth]{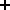}};
        \node (im1_updn) at (6.5,2) {\includegraphics[width=.06\textwidth]{figures/upsample_ex_im1.png}};
        
        \node (im2) at (0,-2) {\includegraphics[width=.06\textwidth]{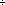}};
        \node (im2_up) at (3.5,-2) {\includegraphics[width=.12\textwidth]{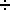}};
        \node (im2_updn) at (6.5,-2) {\includegraphics[width=.06\textwidth]{figures/upsample_ex_im2.png}};
        
        \draw[->] (im1) to node[above, sloped]{downsample} (im1_dn);
        \draw[->] (im1) to node[above]{upsample} (im1_up);
        \draw[->] (im1_dn) to node[above]{upsample} (im1_dnup);
        \draw[->] (im1_up) to node[above]{downsample} (im1_updn);

        \draw[->] (im2) to node[above, sloped]{downsample} (im1_dn);
        \draw[->] (im2) to node[above]{upsample} (im2_up);
        \draw[->] (im2_up) to node[above]{downsample} (im2_updn);
        
        \draw[draw=none] (im2_updn) to node[sloped] {$\neq$} (im1_dnup);
    \end{tikzpicture}
    \caption{Example illustrating the problem with upsampling in a scale-equivariant architecture. We have images $f_1$ and $f_2$ such that when both are downsampled and then upsampled they yield the same result, but if both are upsampled and then downsampled they yield different results.}
    \label{fig:upsample_counter_ex}
\end{figure}

Given any $k\in \mathbb{N}$, take $l=k$, $x = (0, 0)$ and any two lifted images $f_1$ and $f_2$ that coincide on certain points, 
$$f_1(s,\gamma^k y) = f_2(s, \gamma^k y) \quad \forall s \in \mathcal{S}, y \in \Z^2,$$
and are different elsewhere, as illustrated in Figure \ref{fig:upsample_counter_ex}. Let us show that $U_{\gamma^l} R_{\gamma^k, x} f_i \neq R_{\gamma^k, \gamma^l x} U_{\gamma^l} f_i$ either for $i=1$ or $i=2$ or both. The set of points where $f_1$ and $f_2$ coincide implies in particular that  $R_{\gamma^k,0} f_1 = R_{\gamma^k,0} f_2$.
Then we have 
\begin{equation*}
    R_{\gamma^k, 0} U_{\gamma^k} f_1 \neq R_{\gamma^k, 0} U_{\gamma^k} f_2,
\end{equation*}
as $R_{\gamma^k, 0} U_{\gamma^k} f_i : (s, y) \mapsto f_i(\gamma^k s, y)$, and $f_1(\gamma^k s, y) \neq f_2(\gamma^k s, y)$ for $y\notin k\Z^2$. Note that $R_{\gamma^k, 0} U_{\gamma^k}$ is nothing else than an upsampling followed by a downsampling, as in Figure~\ref{fig:upsample_counter_ex}.\\
Since, on the other hand, $R_{\gamma^k,0} f_1 = R_{\gamma^k,0} f_2$, we get
\begin{equation*}
    U_{\gamma^k} R_{\gamma^k, 0} f_1 = U_{\gamma^k} R_{\gamma^k, 0} f_2 .
\end{equation*}
Hence, either $R_{k,0} U_k f_1 \neq U_k R_{k,0} f_1$ or $R_{k,0} U_k f_2 \neq U_k R_{k,0} f_2$ or both, proving our point. 

\section{Proof of Proposition 1}
\begin{proof}
First, consider $k < m$
\begin{align*}
    R_{\gamma^{k}, 0} \circ U_{\gamma^{k}} f (\gamma^p, y) & = (U_{\gamma^k} f) (\gamma^k \gamma^p, \gamma^k y) \\
                                     & = f(\gamma^k \gamma^p, y)
\end{align*}
so $R_{\gamma^k, 0} U_{\gamma^m} f(\gamma^p, y) = U_{\gamma^{m-k}} f(\gamma^{p+m}, y)$.

Now, let $f = U_{\gamma^{m}} f_i \in F$, $k \leq \min \{ n_i | i = 1, \dots, N\} \leq m$, we have
\begin{align*}
    U_{\gamma} R_{\gamma^k,x} f(\gamma^p, y) & = U_{\gamma} R_{\gamma^k,x} U_{\gamma^{m}} f_i(\gamma^p, y) \\
                               & = U_{\gamma} R_{1,x} R_{\gamma^k,0} U_{\gamma^m} f_i(\gamma^p, y) \\
                               & = R_{1,\gamma x} U_{\gamma} U_{\gamma^{m-k}} f_i(\gamma^{p+m} r, y) \\
                               & = U_{\gamma^{m-k+1}} f_i(\gamma^{m+p}, y + \gamma x)
\end{align*}
and, on the other hand
\begin{align*}
    R_{\gamma^k,\gamma x} U_{\gamma} f(\gamma^p, y) & = R_{\gamma^k,\gamma x} U_\gamma  U_{\gamma^m} f_i(\gamma^p, y) \\
                           & = R_{1,\gamma x} R_{\gamma^k ,0} U_{\gamma^{m+1}} f_i(\gamma^p, y) \\
                           & = R_{1,\gamma x} U_{\gamma^{m-k+1}} f_i(\gamma^{p+k}, y) \\
                           & = U_{\gamma^{m-k+1}} f_i(\gamma^{p+m}, y + \gamma x) \\ 
                           & = U_\gamma R_{\gamma^k, x} f(\gamma^p, y),
\end{align*}
implying $U_\gamma R_{\gamma^k, x} f = R_{\gamma^k, \gamma x} U_\gamma f$.
Repeated application gives us the desired result. 
\end{proof}

\section{Details of the Experiments}
\subsection{Oxford IIIT Pet}
All models, except for the augmented U-Net are trained for $300$ epochs. The augmented U-Net is trained for four times as many epochs.
To train all models we apply data augmentation consisting of, rotations by a uniformly sampled angles in $[-10^\circ, 10^\circ]$, linear contrast changes by values in the range $[0.9, 1.1]$, random horizontal flipping and random cropping to size $112 \times 112$.
Learning rate starts at $10^{-3}$ an is reduced by $10$ when the validation loss does not improve for 30 epochs. We use a batch size of $8$.

\subsection{DIC-C2DH-HeLa}
All models, except for the U-Net with jittering $4$ are trained for $200$ epochs. The augmented U-Net with jittering $4$ is trained for four times as many epochs.
To train all models we apply data augmentation consisting of, rotations by a uniformly sampled angles in $[-10^\circ, 10^\circ]$, linear contrast changes by values in the range $[0.9, 1.1]$, random horizontal and vertical flipping and elastic transformations.
Learning rate starts at $10^{-3}$, weight decay starts at $10^{-4}$ and both are reduced by exponential decay such that they are divided by $10$ every $100$ epochs (the decay stops at epoch $300$ for the U-Net with size $4$ jittering). We use a batch size of $1$.

\section{More Examples of Predictions}

\begin{figure}[th]
    \centering
    \begin{minipage}[t]{.22\textwidth}
    \centering
    \includegraphics[width=.125\textwidth]{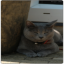}
    \includegraphics[width=.25\textwidth]{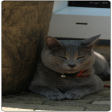}
    \includegraphics[width=.5\textwidth]{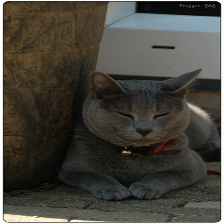}
    
    \includegraphics[width=1.\textwidth]{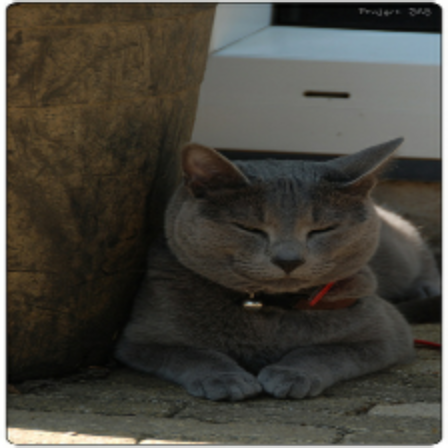}
    
    \end{minipage}
    \begin{minipage}[t]{.22\textwidth}
    \centering
    \includegraphics[width=.125\textwidth]{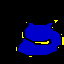}
    \includegraphics[width=.25\textwidth]{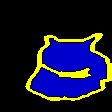}
    \includegraphics[width=.5\textwidth]{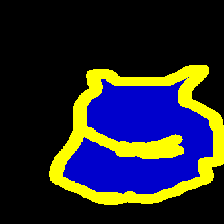}
    
    \includegraphics[width=1.\textwidth]{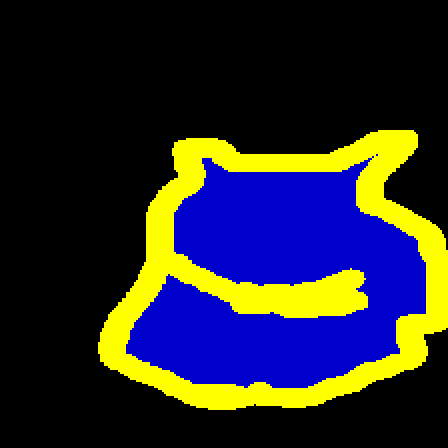}
    
    \end{minipage}
    ~
    \begin{minipage}[t]{.22\textwidth}
    \centering
    \includegraphics[width=.125\textwidth]{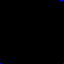}
    \includegraphics[width=.25\textwidth]{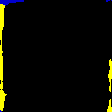}
    \includegraphics[width=.5\textwidth]{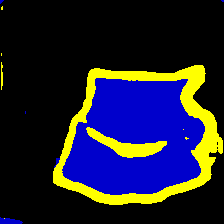}
    
    \includegraphics[width=1.\textwidth]{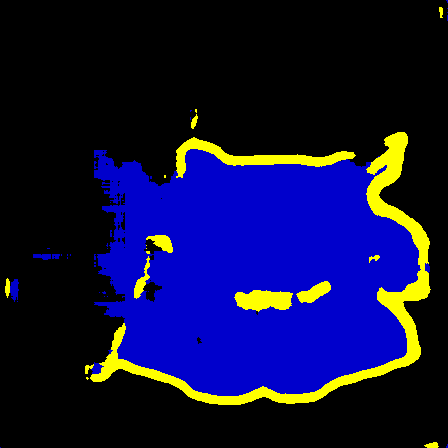}
    
    \end{minipage}
    \begin{minipage}[t]{.22\textwidth}
    \centering
    \includegraphics[width=.125\textwidth]{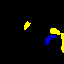}
    \includegraphics[width=.25\textwidth]{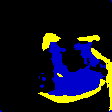}
    \includegraphics[width=.5\textwidth]{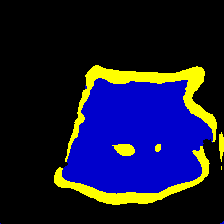}
    
    \includegraphics[width=1.\textwidth]{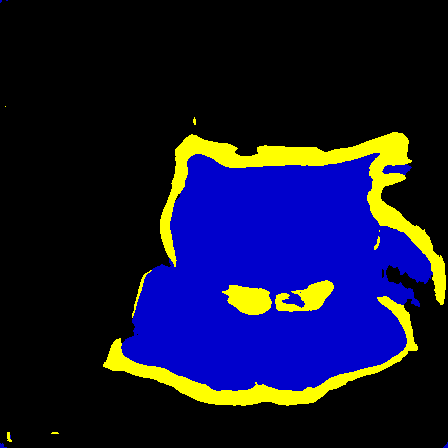}
    
    \end{minipage}
    \begin{minipage}[t]{.22\textwidth}
    \centering
    \includegraphics[width=.125\textwidth]{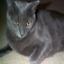}
    \includegraphics[width=.25\textwidth]{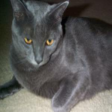}
    \includegraphics[width=.5\textwidth]{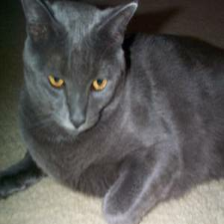}
    
    \includegraphics[width=1.\textwidth]{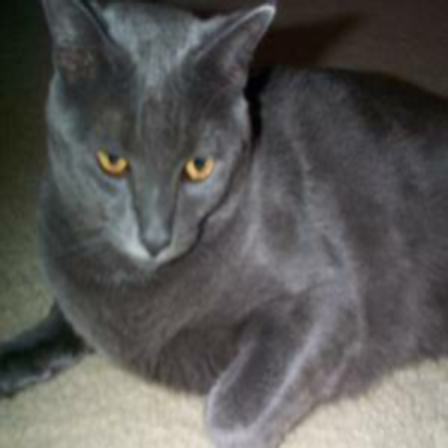}
    
    \end{minipage}
    \begin{minipage}[t]{.22\textwidth}
    \centering
    \includegraphics[width=.125\textwidth]{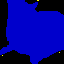}
    \includegraphics[width=.25\textwidth]{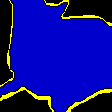}
    \includegraphics[width=.5\textwidth]{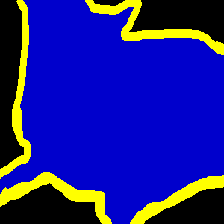}
    
    \includegraphics[width=1.\textwidth]{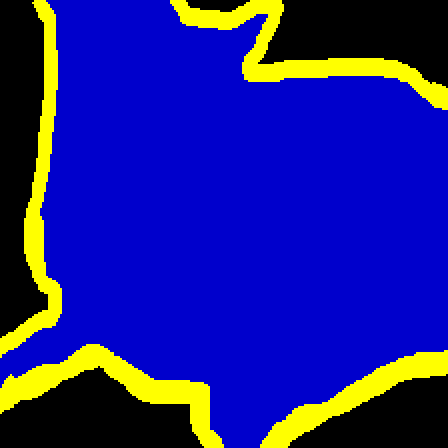}
    
    \end{minipage}
    ~
    \begin{minipage}[t]{.22\textwidth}
    \centering
    \includegraphics[width=.125\textwidth]{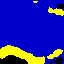}
    \includegraphics[width=.25\textwidth]{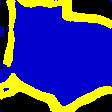}
    \includegraphics[width=.5\textwidth]{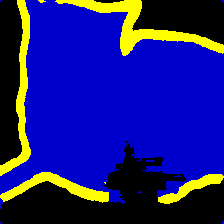}
    
    \includegraphics[width=1.\textwidth]{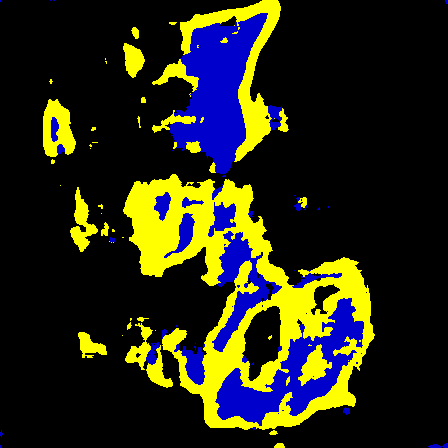}
    
    \end{minipage}
    \begin{minipage}[t]{.22\textwidth}
    \centering
    \includegraphics[width=.125\textwidth]{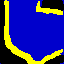}
    \includegraphics[width=.25\textwidth]{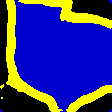}
    \includegraphics[width=.5\textwidth]{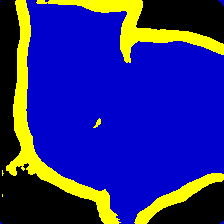}
    
    \includegraphics[width=1.\textwidth]{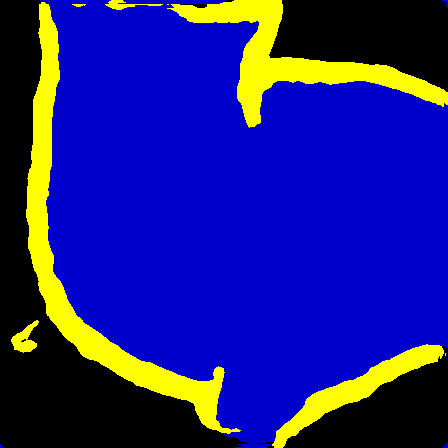}
    
    \end{minipage}
    
    \begin{minipage}[t]{.22\textwidth}
    \centering
    \includegraphics[width=.125\textwidth]{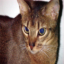}
    \includegraphics[width=.25\textwidth]{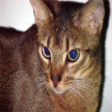}
    \includegraphics[width=.5\textwidth]{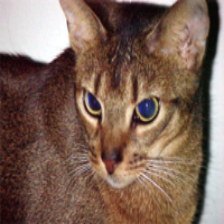}
    
    \includegraphics[width=1.\textwidth]{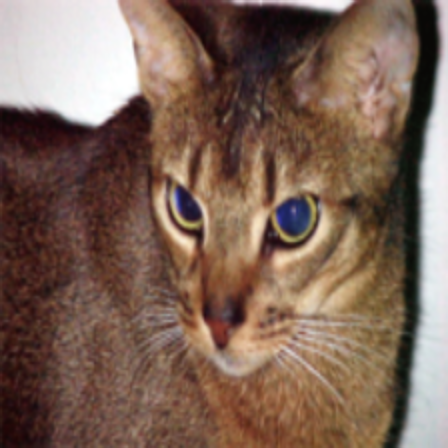}
    
    (a) Image
    \end{minipage}
    \begin{minipage}[t]{.22\textwidth}
    \centering
    \includegraphics[width=.125\textwidth]{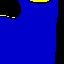}
    \includegraphics[width=.25\textwidth]{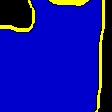}
    \includegraphics[width=.5\textwidth]{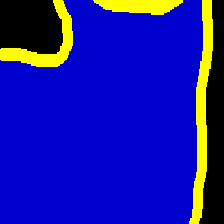}
    
    \includegraphics[width=1.\textwidth]{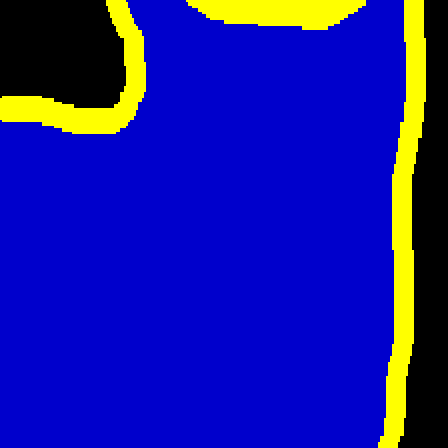}
    
    (b) Ground Truth
    \end{minipage}
    ~
    \begin{minipage}[t]{.22\textwidth}
    \centering
    \includegraphics[width=.125\textwidth]{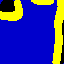}
    \includegraphics[width=.25\textwidth]{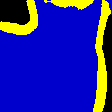}
    \includegraphics[width=.5\textwidth]{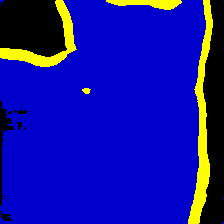}
    
    \includegraphics[width=1.\textwidth]{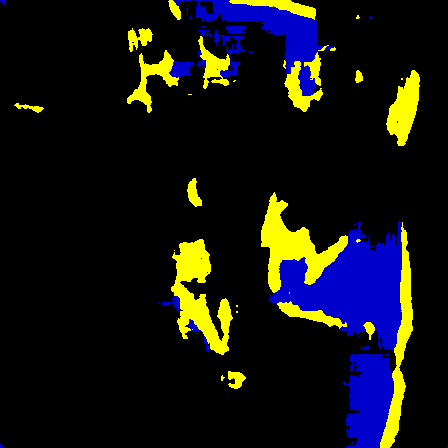}
    
    (c) U-Net
    \end{minipage}
    \begin{minipage}[t]{.22\textwidth}
    \centering
    \includegraphics[width=.125\textwidth]{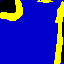}
    \includegraphics[width=.25\textwidth]{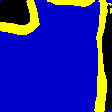}
    \includegraphics[width=.5\textwidth]{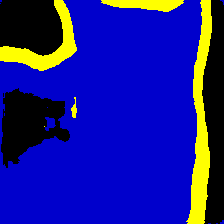}
    
    \includegraphics[width=1.\textwidth]{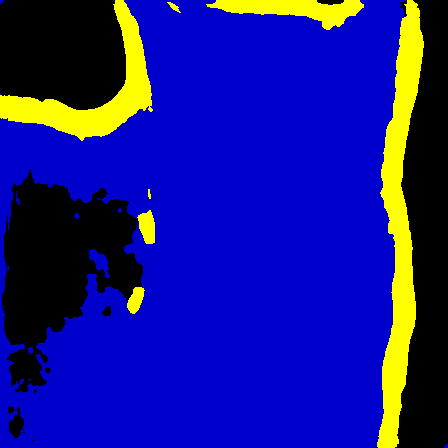}
    
    (d) SEU-Net
    \end{minipage}
    \caption{Sample test images at different scales and ground truth from the Oxford-IIIT Pet dataset, along with the U-Net and SEU-Net predictions. The scales present are $0.25, 0.5, 1$ and $2$ times the training scale.}
    \label{fig:example_pets2}
\end{figure}

\begin{figure}[th]
    \centering
    \begin{minipage}[t]{.22\textwidth}
    \centering
    \includegraphics[width=.2\textwidth]{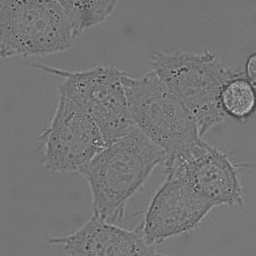}
    \includegraphics[width=.4\textwidth]{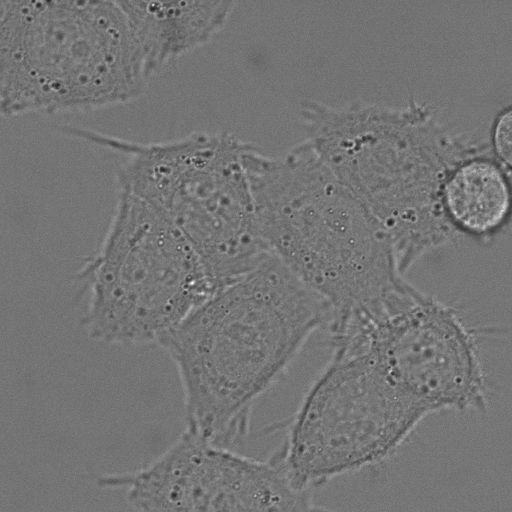}
    
    \includegraphics[width=.8\textwidth]{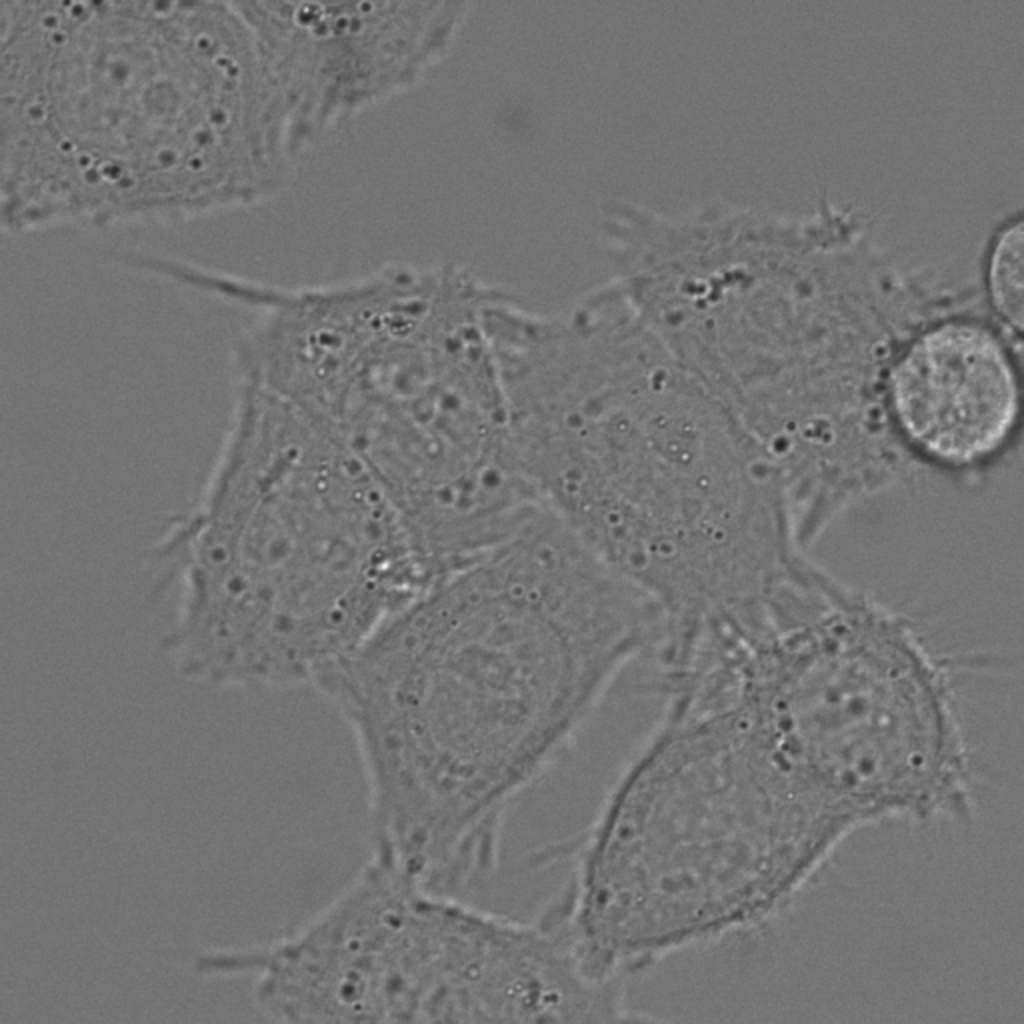}
    
    \end{minipage}
    \begin{minipage}[t]{.22\textwidth}
    \centering
    \includegraphics[width=.2\textwidth]{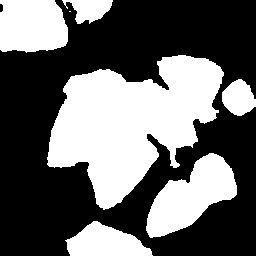}
    \includegraphics[width=.4\textwidth]{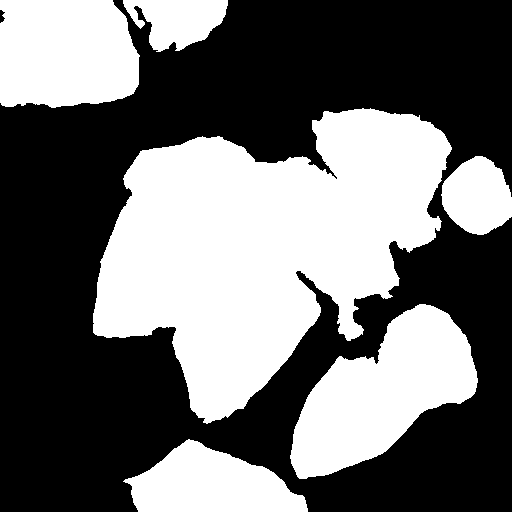}
    
    \includegraphics[width=.8\textwidth]{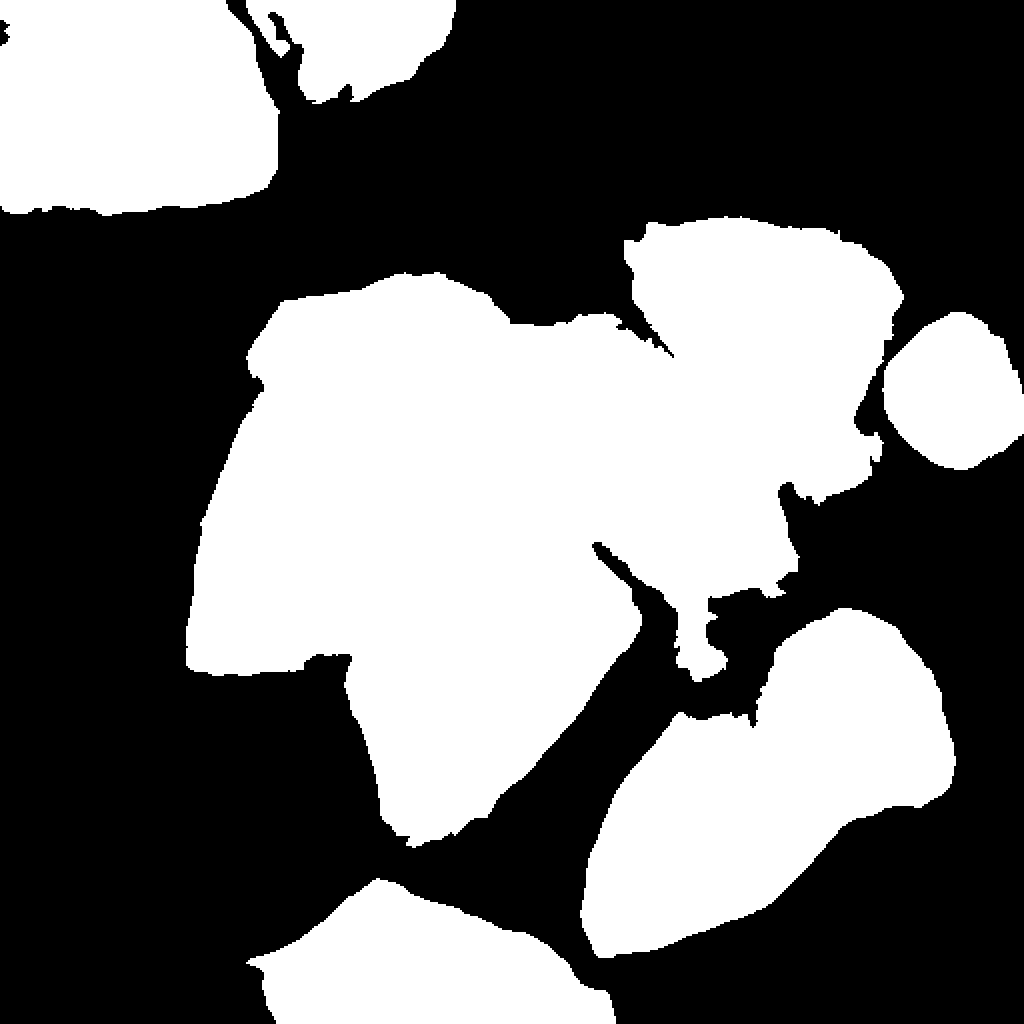}
    
    \end{minipage}
    ~
    \begin{minipage}[t]{.22\textwidth}
    \centering
    \includegraphics[width=.2\textwidth]{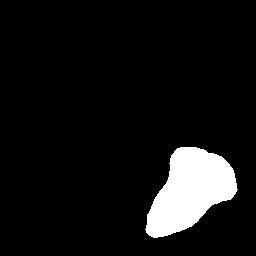}
    \includegraphics[width=.4\textwidth]{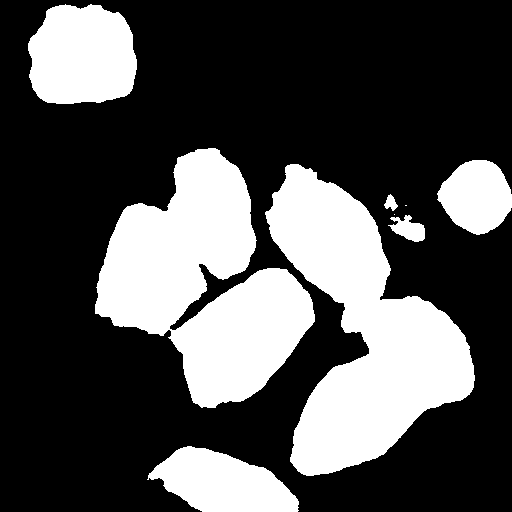}
    
    \includegraphics[width=.8\textwidth]{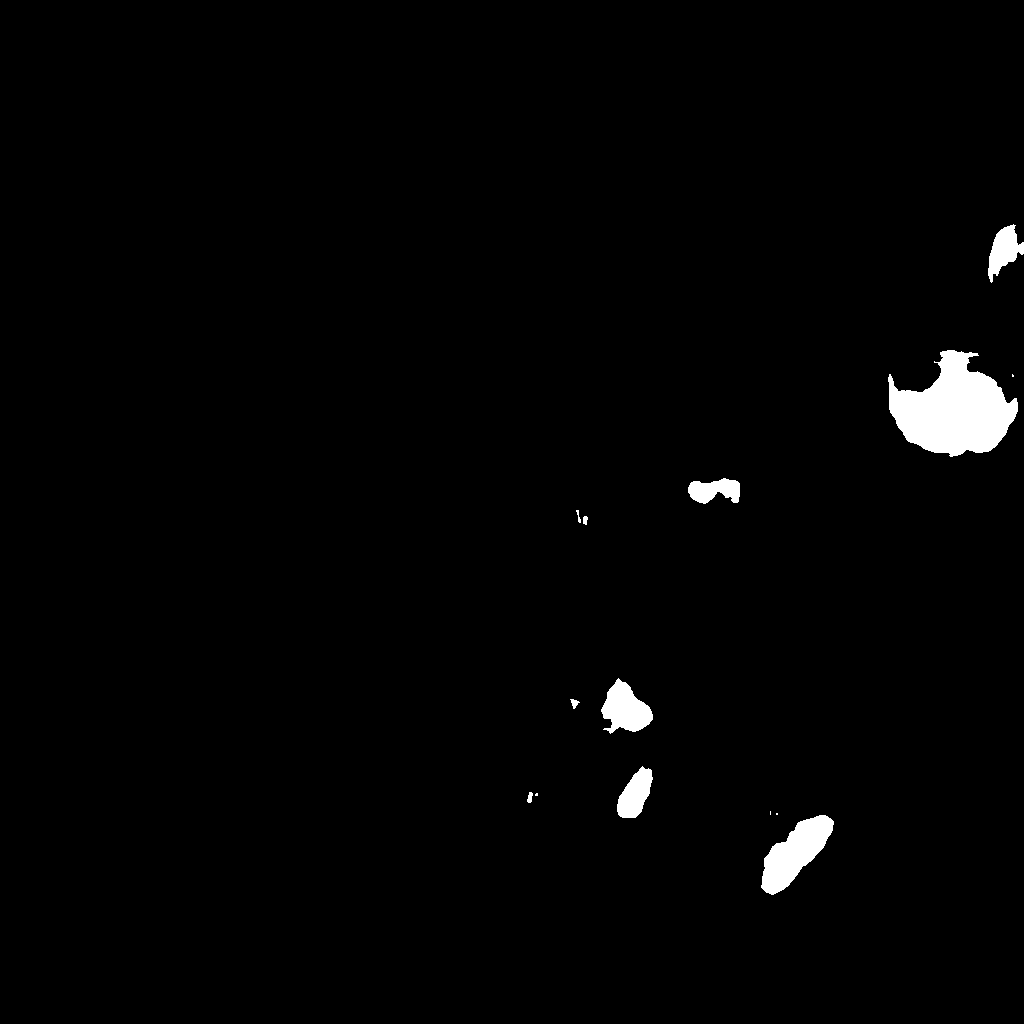}
    
    \end{minipage}
    \begin{minipage}[t]{.22\textwidth}
    \centering
    \includegraphics[width=.2\textwidth]{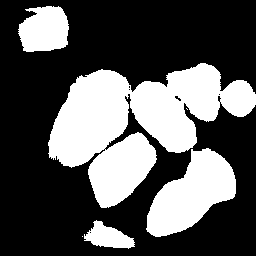}
    \includegraphics[width=.4\textwidth]{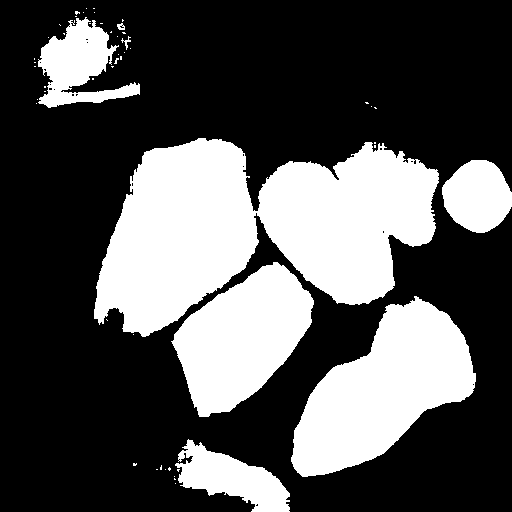}
    
    \includegraphics[width=.8\textwidth]{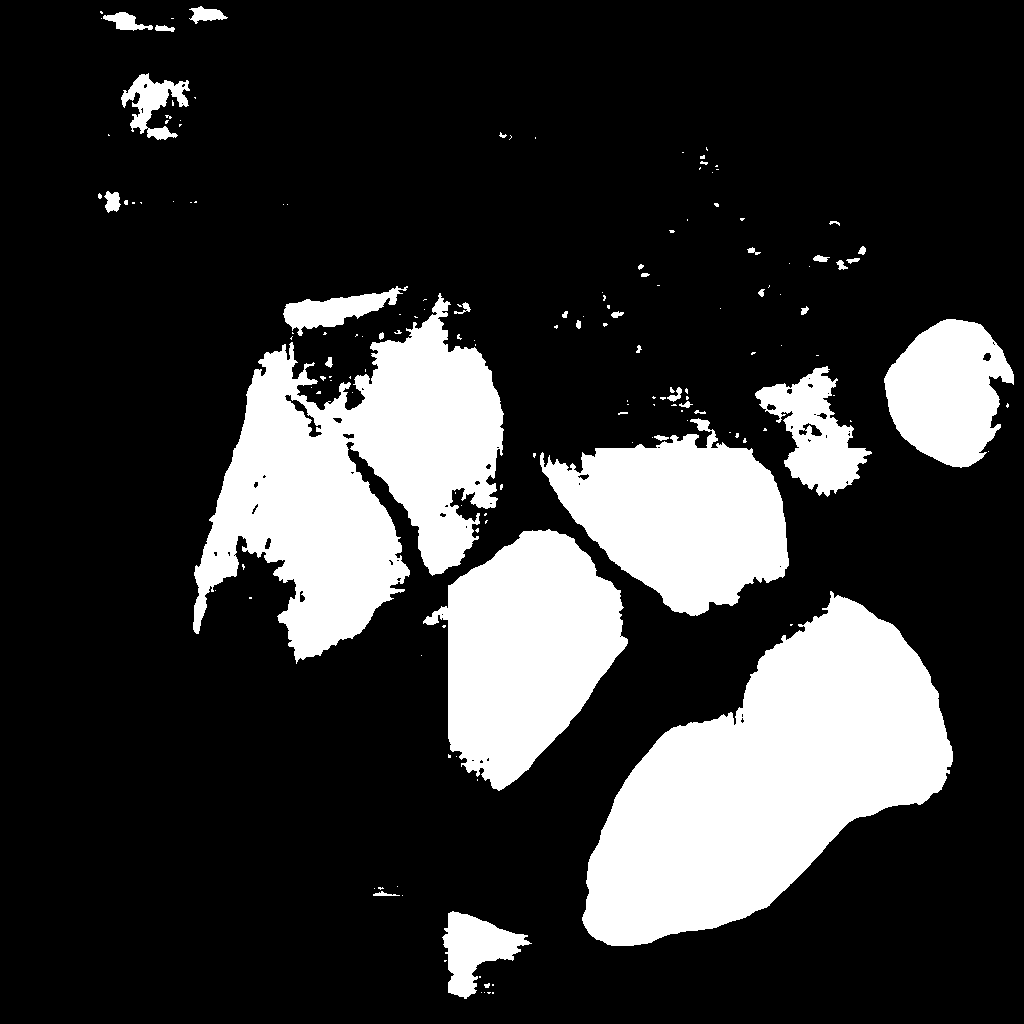}
    
    \end{minipage}
    \begin{minipage}[t]{.22\textwidth}
    \centering
    \includegraphics[width=.2\textwidth]{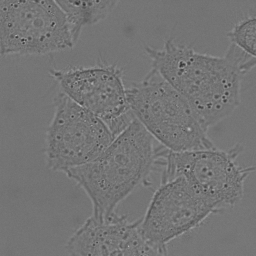}
    \includegraphics[width=.4\textwidth]{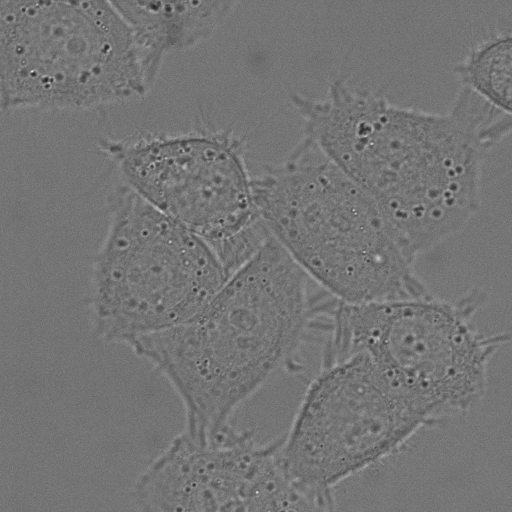}
    
    \includegraphics[width=.8\textwidth]{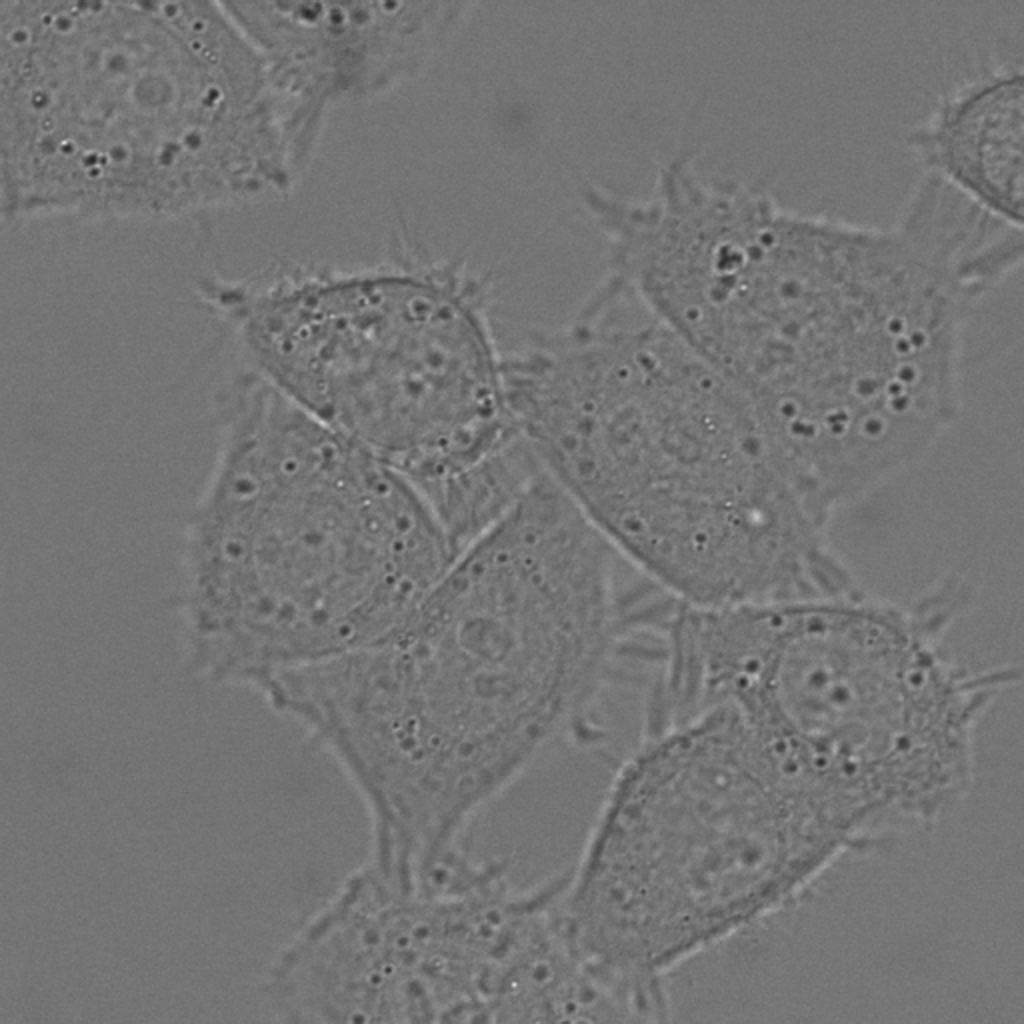}
    
    \end{minipage}
    \begin{minipage}[t]{.22\textwidth}
    \centering
    \includegraphics[width=.2\textwidth]{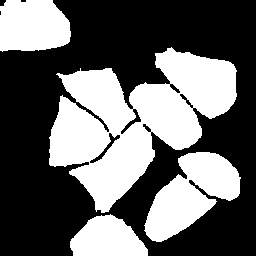}
    \includegraphics[width=.4\textwidth]{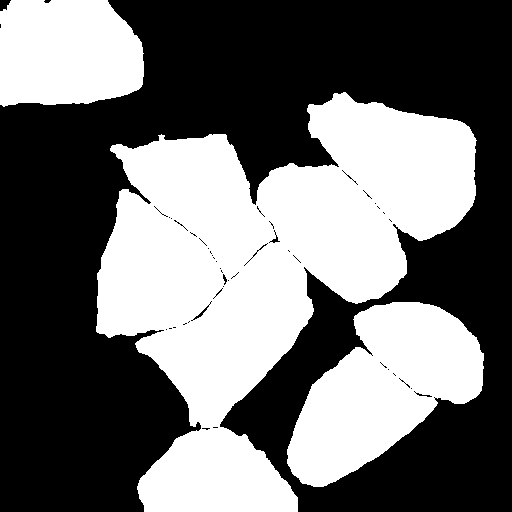}
    
    \includegraphics[width=.8\textwidth]{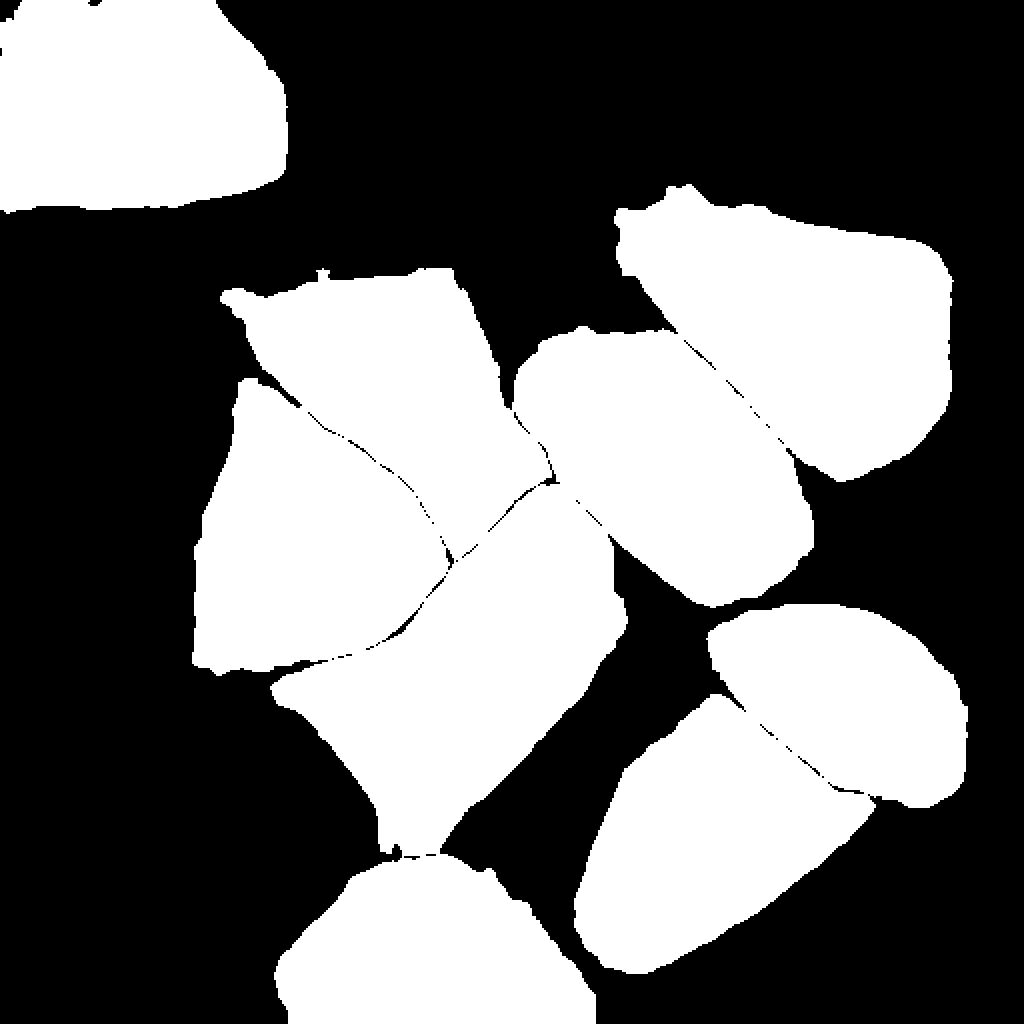}
    
    \end{minipage}
    ~
    \begin{minipage}[t]{.22\textwidth}
    \centering
    \includegraphics[width=.2\textwidth]{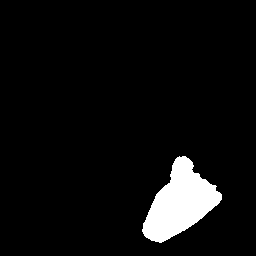}
    \includegraphics[width=.4\textwidth]{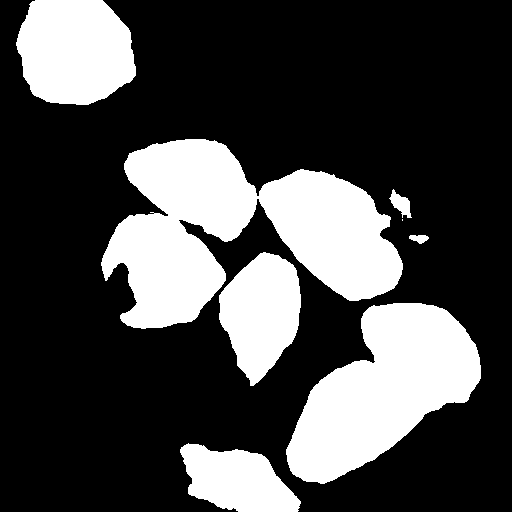}
    
    \includegraphics[width=.8\textwidth]{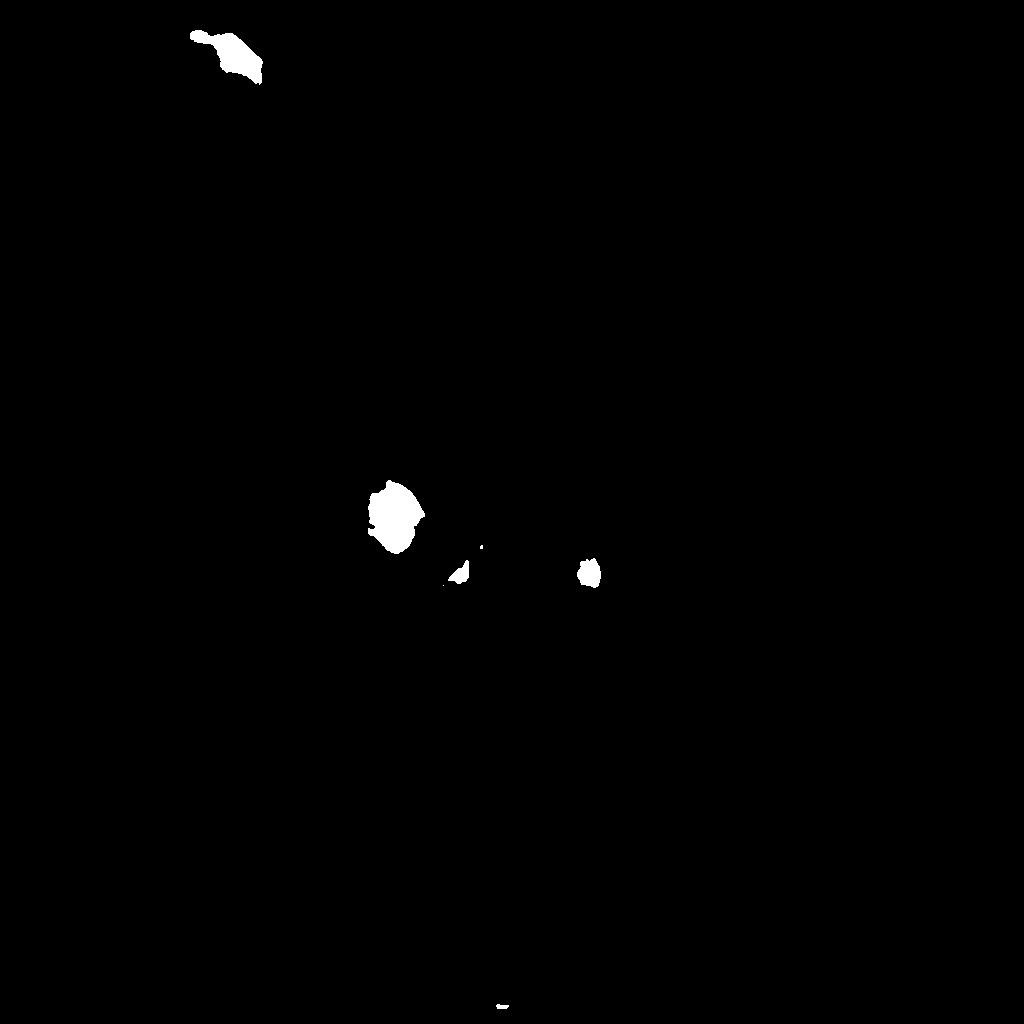}
    
    \end{minipage}
    \begin{minipage}[t]{.22\textwidth}
    \centering
    \includegraphics[width=.2\textwidth]{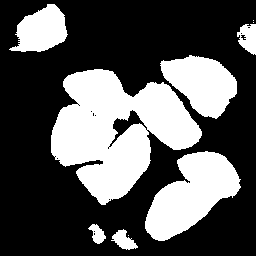}
    \includegraphics[width=.4\textwidth]{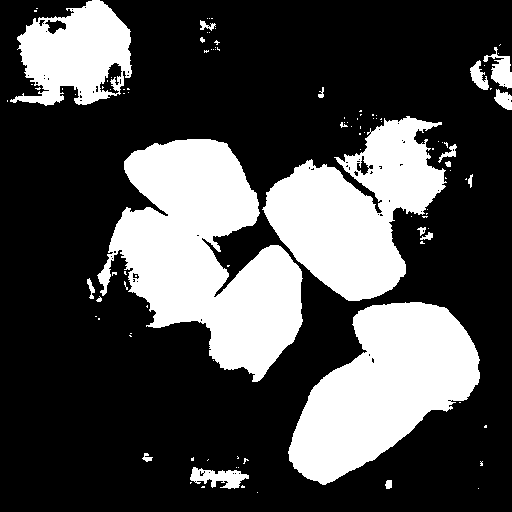}
    
    \includegraphics[width=.8\textwidth]{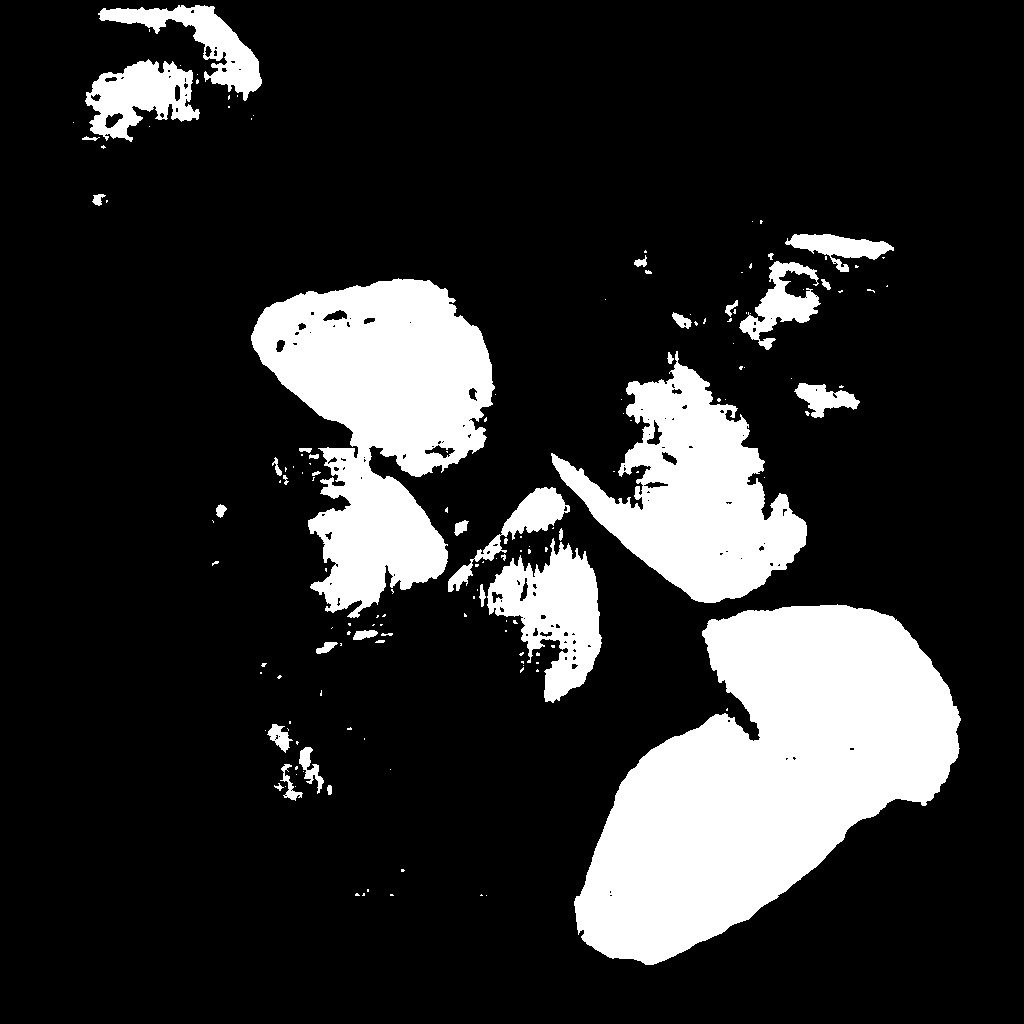}
    
    \end{minipage}
    
    \begin{minipage}[t]{.22\textwidth}
    \centering
    \includegraphics[width=.2\textwidth]{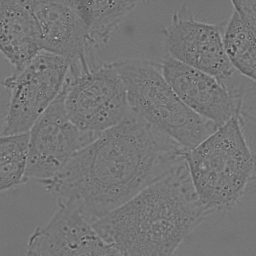}
    \includegraphics[width=.4\textwidth]{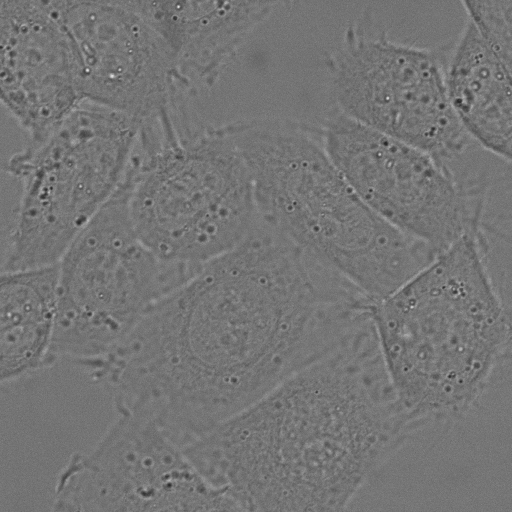}
    
    \includegraphics[width=.8\textwidth]{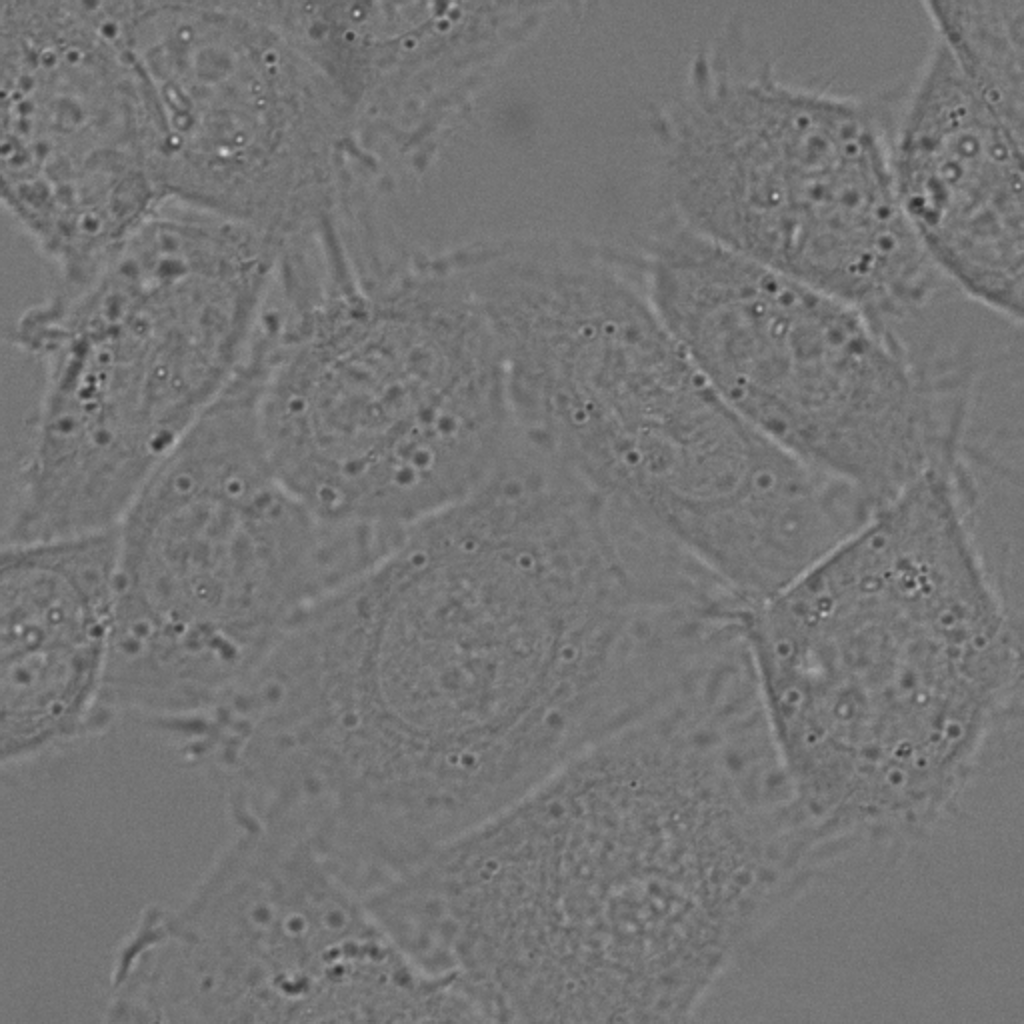}
    
    (a) Image
    \end{minipage}
    \begin{minipage}[t]{.22\textwidth}
    \centering
    \includegraphics[width=.2\textwidth]{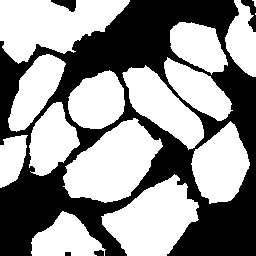}
    \includegraphics[width=.4\textwidth]{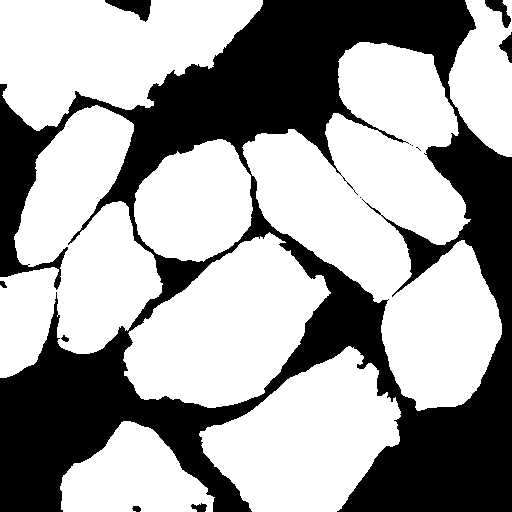}
    
    \includegraphics[width=.8\textwidth]{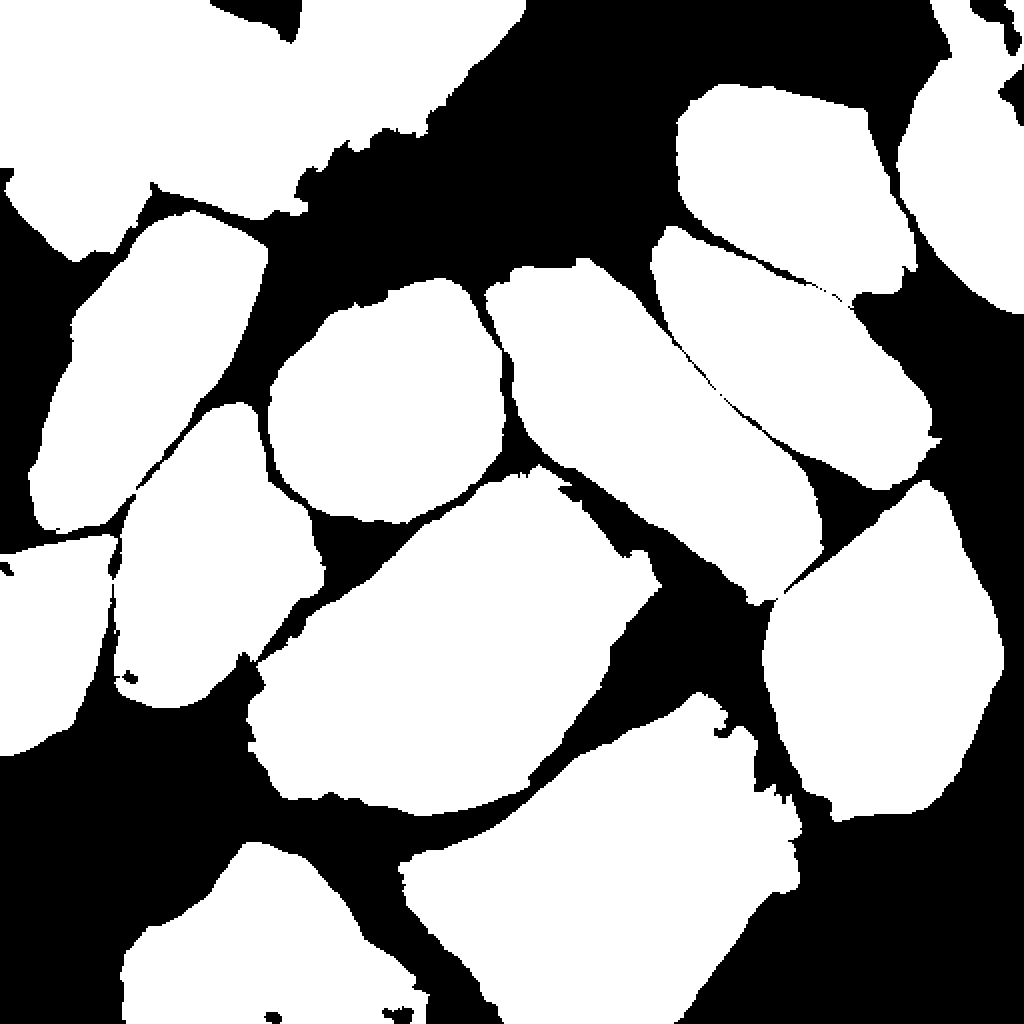}
    
    (b) Ground Truth
    \end{minipage}
    ~
    \begin{minipage}[t]{.22\textwidth}
    \centering
    \includegraphics[width=.2\textwidth]{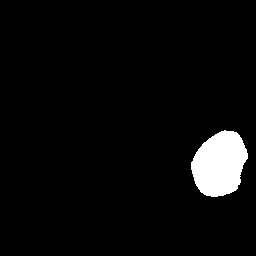}
    \includegraphics[width=.4\textwidth]{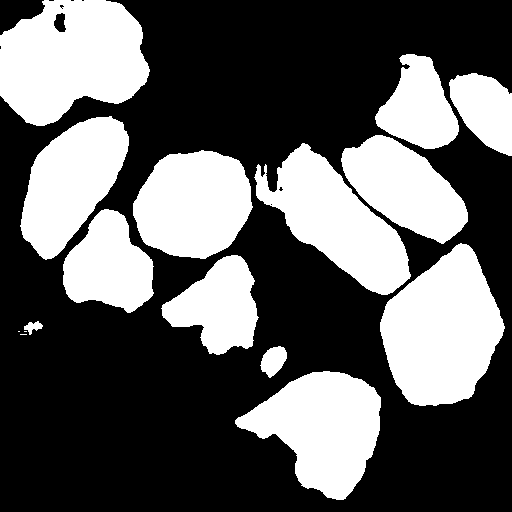}
    
    \includegraphics[width=.8\textwidth]{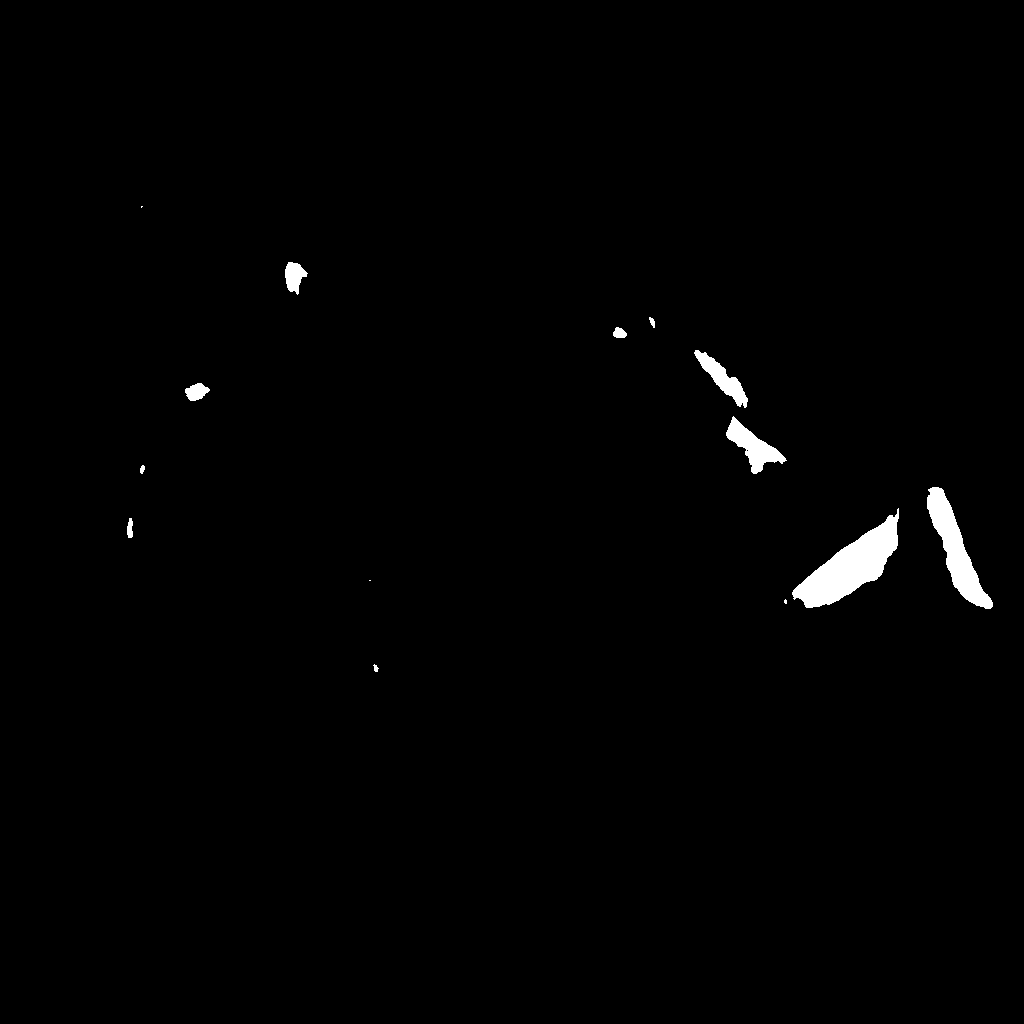}
    
    (c) U-Net
    \end{minipage}
    \begin{minipage}[t]{.22\textwidth}
    \centering
    \includegraphics[width=.2\textwidth]{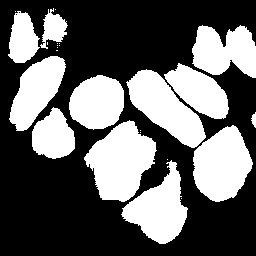}
    \includegraphics[width=.4\textwidth]{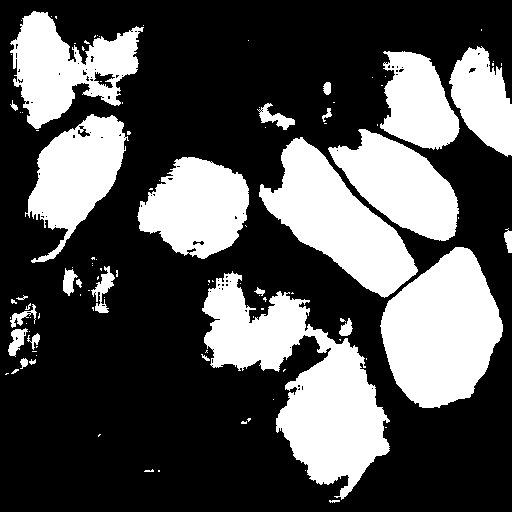}
    
    \includegraphics[width=.8\textwidth]{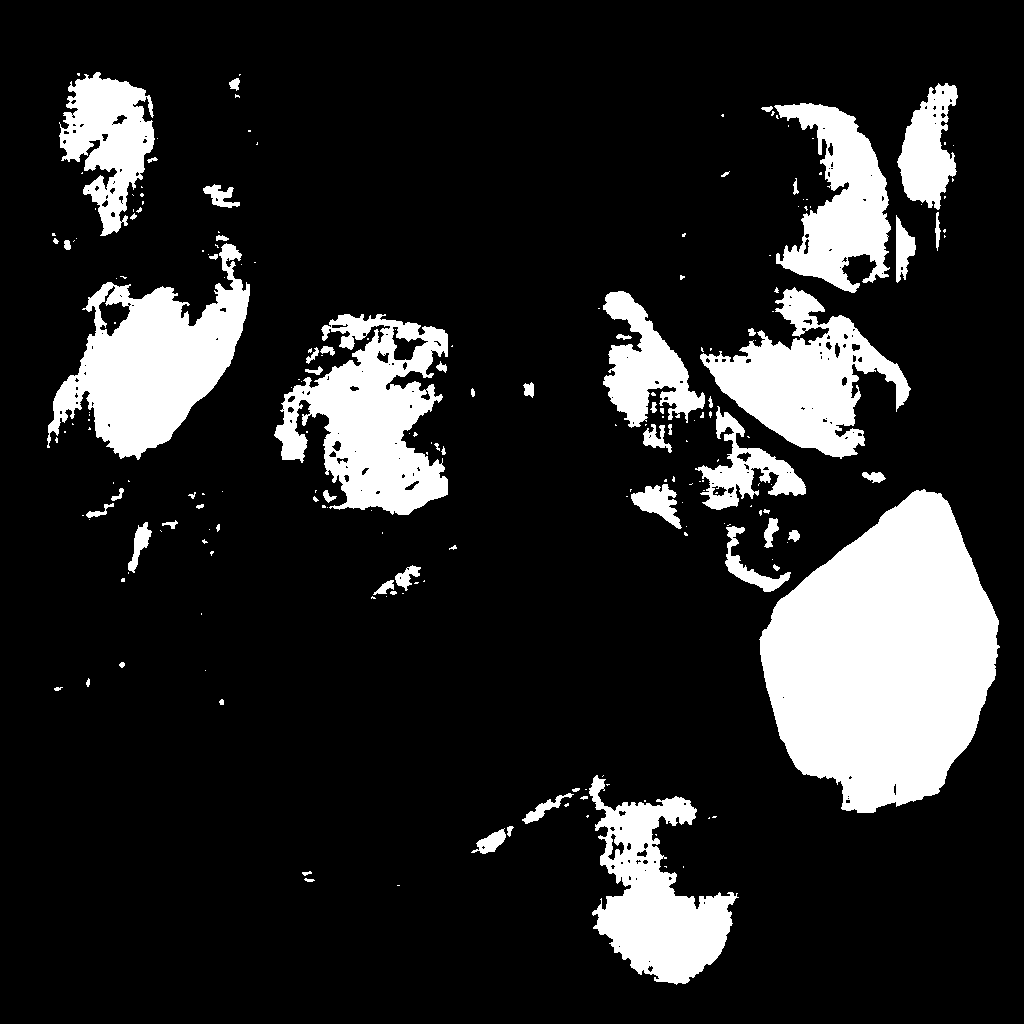}
    
    (d) SEU-Net
    \end{minipage}
    \caption{Predictions from DIC-HeLa at different scales, namely scales $0.5$, $1$ and $2$.}
    \label{fig:example_cells2}
\end{figure}

In Figure \ref{fig:example_pets2} we can see some more examples of predictions from the Oxford Pet dataset, particularly when the U-Net struggles to generalize to new scales. Similarly Figure \ref{fig:example_cells2} showcases some extra examples from the experiment from the DIC-HeLa experiment.

\end{document}